\documentclass[pdflatex,sn-mathphys-num]{sn-jnl}

\usepackage{graphicx}%
\usepackage{multirow}%
\usepackage{amsmath,amssymb,amsfonts}%
\usepackage{amsthm}%
\usepackage{mathrsfs}%
\usepackage[title]{appendix}%
\usepackage{xcolor}%
\usepackage{textcomp}%
\usepackage{manyfoot}%
\usepackage{booktabs}%
\usepackage{algorithm}%
\usepackage{algorithmicx}%
\usepackage{algpseudocode}%
\usepackage{listings}%

\usepackage{subfigure}
\usepackage{xurl}

\theoremstyle{thmstyleone}%
\theoremstyle{thmstyletwo}%

\theoremstyle{thmstylethree}%

\begin{document}

\title[Benchmarking Deep Reinforcement Learning for Navigation in Denied Sensor Environments]{Benchmarking Deep Reinforcement Learning for Navigation in Denied Sensor Environments}

\author*[]{\fnm{Mariusz} \sur{Wisniewski}}\email{m.wisniewski@cranfield.ac.uk}

\author[]{\fnm{Paraskevas} \sur{Chatzithanos}}

\author*[]{\fnm{Weisi} \sur{Guo}}

\author[]{\fnm{Antonios} \sur{Tsourdos}}

\affil{\orgdiv{Centre for Autonomous and Cyberphysical Systems}, \orgname{Cranfield University}, \orgaddress{\street{College Road}, \city{Cranfield}, \postcode{MK43 0AL}, \state{Bedfordshire}, \country{UK}}}

\abstract{Deep Reinforcement learning (DRL) is used to enable autonomous navigation in unknown environments. Most research assume perfect sensor data, but real-world environments may contain natural and artificial sensor noise and denial. Here, we present a benchmark of both well-used and emerging DRL algorithms in a navigation task with configurable sensor denial effects. In particular, we are interested in comparing how different DRL methods (e.g. model-free PPO vs. model-based DreamerV3) are affected by sensor denial. We show that DreamerV3 outperforms other methods in the visual end-to-end navigation task with a dynamic goal - and other methods are not able to learn this. Furthermore, DreamerV3 generally outperforms other methods in sensor-denied environments. In order to improve robustness, we use adversarial training and demonstrate an improved performance in denied environments, although this generally comes with a performance cost on the vanilla environments. We anticipate this benchmark of different DRL methods and the usage of adversarial training to be a starting point for the development of more elaborate navigation strategies that are capable of dealing with uncertain and denied sensor readings. 
}


\maketitle

\section{Introduction}\label{section:Introduction}

Autonomous navigation is a fundamental challenge in unmanned systems. 
Conventional autonomous systems typically use Simultaneous Localisation and Mapping (SLAM) alongside Kalman filters that fuse data from multiple sources to map the environment, localize the agent, and generate trajectories to the goal. 
In recent years, deep reinforcement learning (DRL) methods have been used to train end-to-end policies that can navigate using a camera.
These improvements have been made aided by advances in digital image processing largely due to convolutional neural networks trained using backpropagation \cite{krizhevsky_imagenet_2012, deng_imagenet_2009, he_deep_2015}, and parametrization of RL using deep neural networks \cite{mnih_human-level_2015}. 

Whilst autonomous visual navigation is achievable in simulated environments \cite{8742226}, most of the research assumes perfect sensor readings. But, in real-world scenarios, information about the environment may be uncertain, incomplete, or even intentionally denied. This can limit the ability of the navigation system to adapt to changes in the environment and make optimal decisions.
These failure modes and how they affect the systems need to be understood for safety-critical operations, such as unmanned aerial vehicles or autonomous road vehicles. In the case of DRL agents, which are notoriously hard to explain, it is important to study how the sensor artefacts or failures affect the training and evaluation. 

Further, real systems typically use a suite of sensors. Whilst a lot of the cutting-edge work in DRL focuses on end-to-end visual navigation, in practice it is important to understand how the models learn across different modalities, by taking in data from multiple sensors. 
In this work, we study how sensor failure affects the training and evaluation of policies that use a camera or Lidar as the observation space.

To understand the effects of sensor noise and failures, we compare several DRL architectures: TD3, PPO, PPO-LSTM, and DreamerV3. TD3 and PPO are a commonly used for evaluating environments with continuous action spaces, PPO-LSTM is the recurrent version of PPO (recurrent models were shown to outperform non-recurrent counterparts by Mirowski et al. \cite{mirowski_learning_2017}), and DreamerV3 which uses autoencoders to create an internal world model of the environment, and was shown to outperform PPO on certain environments. 

We generate a baseline of their performance in a 'vanilla' environment, with no perturbations - noise or failures - to the sensors. We then progressively add random areas to the environment in which the sensors are perturbed. The policies are then evaluated to understand how successful they are in navigation under varying degrees of noise.
To train and evaluate the methods we use a modified version of the ROS-based gymnasium environment for training navigation policies, \textit{DRL-Robot-Navigation} \cite{cimurs_goal-driven_2021}. It contains a 3D maze in which a robot is tasked with navigating to a goal. This publication is an extension of the conference paper on autonomous navigation presented at ICUAS 2024 \cite{wisniewski_autonomous_2024}. 

\subsection{Review of DRL Applications to Navigation}
Several studies have achieved end-to-end navigation using DRL with the camera as the sole sensor in a simulation environment \cite{zhu_deep_2021}. 
Popular environments include the DeepMind Lab \cite{beattie_deepmind_2016}, ViZDoom \cite{kempka_vizdoom_2016}, and DRL-Robot-Navigation \cite{cimurs_goal-driven_2021}. 

Mirowski et al. \cite{mirowski_learning_2017} present an application of the A3C \cite{mnih_asynchronous_2016} algorithm to the Deepmind Lab environment. They use the camera as the sole sensor and their best-performing architecture consists of the image encoder, 2 LSTM layers, depth and loop predictions.
The agent is rewarded for reaching the goal, or subgoals in the form of a fruit. The agent is not penalized for wall collisions, allowing strategies such as 'wall-following' that do not require memory. 
DRL can also be used to race FPV unmanned aerial vehicles (UAVs) against expert human pilots \cite{kaufmann_champion-level_2023} by using the image sensor combined with the IMU (which are combined to detect gates and generate VIO states input to the DRL algorithm) in static environments. 
Polvara et al. show how a hierarchy of DQNs can be used for the autonomous landing of UAVs \cite{DRL}. Their method handles a large variety of simulated environments and outperforms human pilots in some conditions.
Interpretable deep reinforcement learning methods for end-to-end autonomous driving can handle complex urban scenarios \cite{Int-RL}. Their method outperforms many baselines including Deep Q-Networks (DQNs), deep deterministic policy gradient (DDPG), Twin Delayed DDPG (TD3), and Soft Actor Critic (SAC) in urban scenarios with surrounding vehicles.
The memory of FPV agents in 3D mazes is evaluated by Pasukonis et al. \cite{pasukonis_evaluating_2022}. They find that models such as Dreamer and IMPALA can match humans on smaller mazes (9x9), but their performance significantly drops off on larger mazes (15x15). 
Lample and Chaplot \cite{lample_playing_2018} show that DRL can be used to train agents to play the first-person-shooter game DOOM, in which the agent is required to navigate a 3D maze, collect items, and shoot at enemies. They found that by using DQN and DQRN with an LSTM layer, the agent can converge, but required an addition of predicting game features (using a separate loss function) to be able to reliably shoot at enemies. 
 Liquid neural networks \cite{Liquid} have been proposed in recent years to achieve robust navigation for changing environments \cite{Liquid2}. In most of these cases, whilst the physical environment is unknown or stochastic, the papers did not consider compromised sensor readings (bar liquid neural networks that consider noise and other perturbations).

\subsection{Review of Sensor Fusion under Attacks}
\subsubsection{Trajectory Estimation Approaches}
The first approach deals with intermittent attacks or denial, for example when GNSS coverage is lost and a Kalman filter (or more advanced recurrent neural network) is used to hold over the trajectory estimation until coverage recovers \cite{10156526, Fusion}. This is useful for brief outages. The same idea can preemptively discover other V2V networks to improve localisation \cite{KalmanNetwork}. However, if the environment requires constant sharp manoeuvring to avoid dynamic objects, this scenario rules out trajectory estimation approaches. Also, if the attacks are continuous and erode the holdover accuracy, this would be unsuitable.

\subsubsection{Sensor Fusion and Generative AI Approaches}
The alternative is suitable for high-end platforms that can afford a diverse sensor suite, performing sensor fusion to alleviate the impact of one sensor being compromised. This is usually unsuitable for low-end drones and does not guarantee success \cite{Fusion2}. A cheaper alternative is to use generative AI to generate alternative sensor representations using a single cheap sensor. For example, infrared images can be generated using a camera sensor and then fused together to achieve more robust recognition of the environment \cite{GAN}. However, these approaches require an onboard diverse sensor suite or a powerful GAN operating onboard, which does not guarantee success.

\subsubsection{Adversarial Environmental Perturbations in DRL}
First-person-view (FPV) navigation problems are typically considered partially observable \cite{spaan_partially_2012} because only a portion of the entire state is visible to the agent at any given time. But, sensor faults also have to be considered. 
Robust partially observable MDP \cite{osogami_robust_2015} considers the POMDP to have uncertain parameters.
Other approaches \cite{tessler_action_2019} consider the interaction between the adversary and the agent to be a zero-sum minmax game and define it as a Probabilistic Action Robust MDP.
Zhang et al. \cite{zhang_robust_2020} consider the addition of adversarial perturbations as a state-adversarial MDP (SA-MDP). In this framing, the state returned by the environment may be perturbed by an adversary (an adversary may be another agent or an environmental process that perturbs the state observations), making the observation uncertain and resulting in the action returned by the agent being potentially sub-optimal. 
Because we consider that the adversary can be static (e.g. it is part of the environment), we prefer the SA-MDP framing for our problem.

When building robust RL models, some of these works focus on physical perturbations \cite{pinto_robust_2017} that may affect the physics of the agent (e.g. by introducing uncertain forces), whilst others consider perturbations to the sensor state \cite{korkmaz_adversarial_2021}. Korkmaz \cite{korkmaz_adversarial_2023, korkmaz_understanding_2024} argues that vanilla training results in more robust policies compared with state-of-the-art adversarial and robust training techniques. Havens et al. \cite{havens_online_2018} propose a hierarchical RL method of switching sub-policies in the presence of adversaries.
Applications of adversarial training \cite{panda_action_2024} show how a zero-sum action-robust RL method can be used to reduce conflicts in air mobility traffic management scenarios. 

\subsection{Gap and Novelty}
Although research on DRL and sensor attacks exists, we have not found elaborate studies combining the effect of sensor failure on DRL performance during learning navigation policies. 
We aim to understand the effects of sensor failures - a potentially disastrous failure mode in autonomous systems - on the ability of DRL agents to learn navigation policies. To do this we:
\begin{itemize}
    \item Benchmark different reinforcement learning algorithms on the \textit{DRL-Robot-Navigation} environment.
    \item Study the effect of observation spaces - Lidar and camera - on the navigation capabilities.
    \item Perform quantitative and qualitative studies showing the effects of adversarial perturbations - noise and sensor failure - on the performance of DRL algorithms in the navigation task.  
\end{itemize}
Further, to enable others to replicate our work, we open source the modifications to the \textit{DRL-Robot-Navigation} environment, and update it to fit the gymnasium \cite{towers_gymnasium_2024} interface.\footnote{\url{https://github.com/mazqtpopx/cranfield-navigation-gym}}

In this paper, we explain the methodology, i.e. the environment, changes to the environment, and the choice of reinforcement learning models, in section \ref{section:Methodology}. We present the results of the benchmark of different algorithms on the environment, the Lidar and camera noise studies, and discussion with respect to literature in section \ref{section:Results}. Finally, we conclude and make suggestions for further work in section \ref{section:Conclusion}.

\section{Methodology}\label{section:Methodology}

\begin{figure}[!t]
\centering
\includegraphics[width=\textwidth]{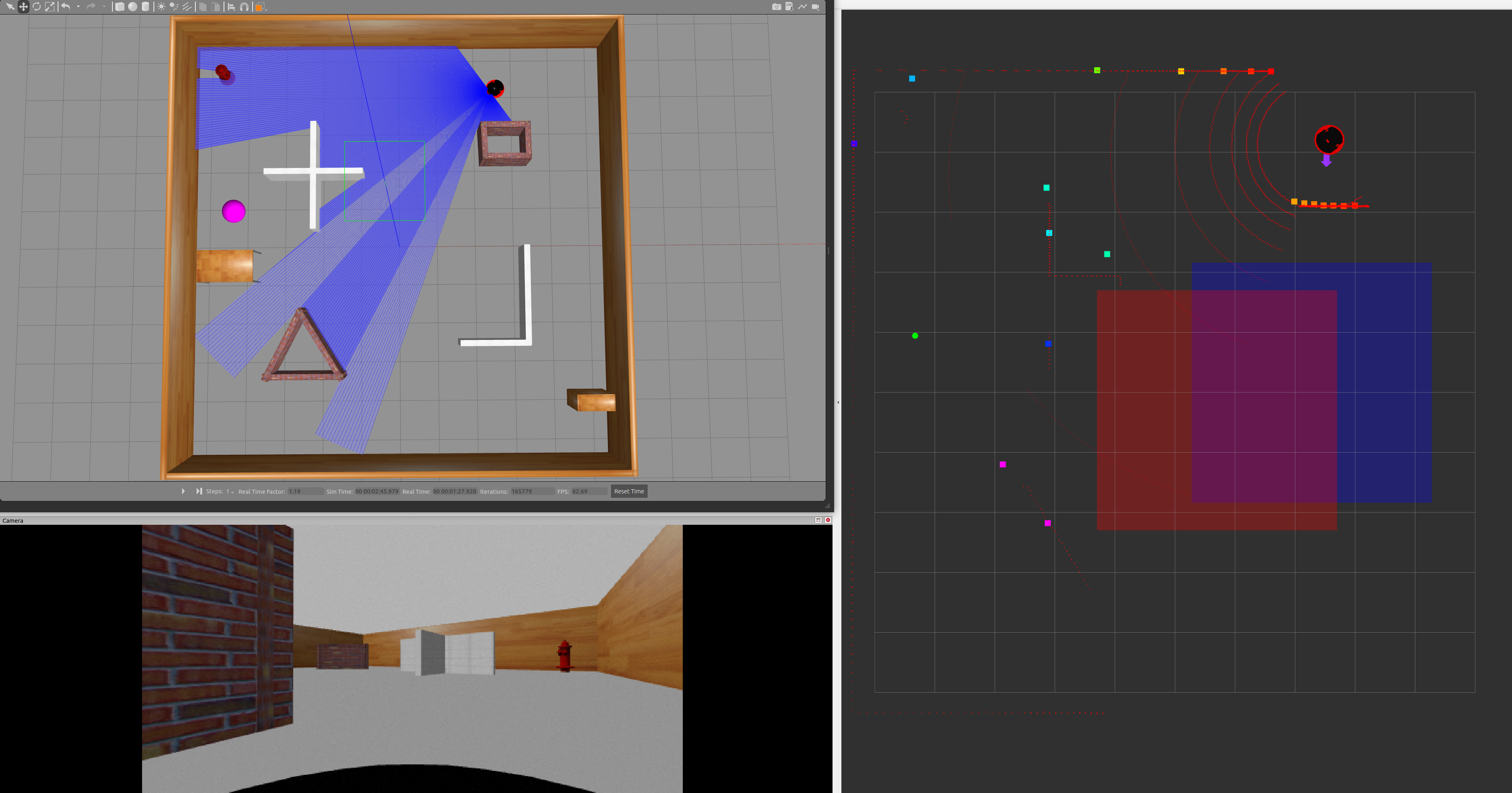}
\caption{Example of the \textit{DRL-Robot-Nav Maze} Environment \cite{cimurs_goal-driven_2021}. Top-left: top-down view of the environment rendered in Gazebo. Bottom-left: the view from the RGB camera. Right: RViz visualization of the discretized Lidar sensor readings, the robot, and the noisy sensor areas (blue: camera noise area, red: Lidar noise area). The pink sphere (top left) is added as a visual cue to mark the goal for vision-based policies.  }
\label{fig:methodology:drl_env}
\end{figure}

We build on top of the \textit{DRL-Robot-Navigation} environment \cite{cimurs_goal-driven_2021} which simulates a robot navigating around a 10 m x 10 m maze, populated with static obstacles. The robot has a suite of sensors - Lidar, camera, and odometry - but only a subset of these was used as the state observation to train the reinforcement learning model in the original paper. The original paper uses Lidar, distance to goal, and angle to goal, concatenated together as the observation space, and the camera is not used. An illustration of the environment, which shows the \textit{DRL-Robot-Navigation} environment along with sensor readings, is presented in Fig. \ref{fig:methodology:drl_env}.

\begin{figure}[h]
\centering
\includegraphics[width=\textwidth]{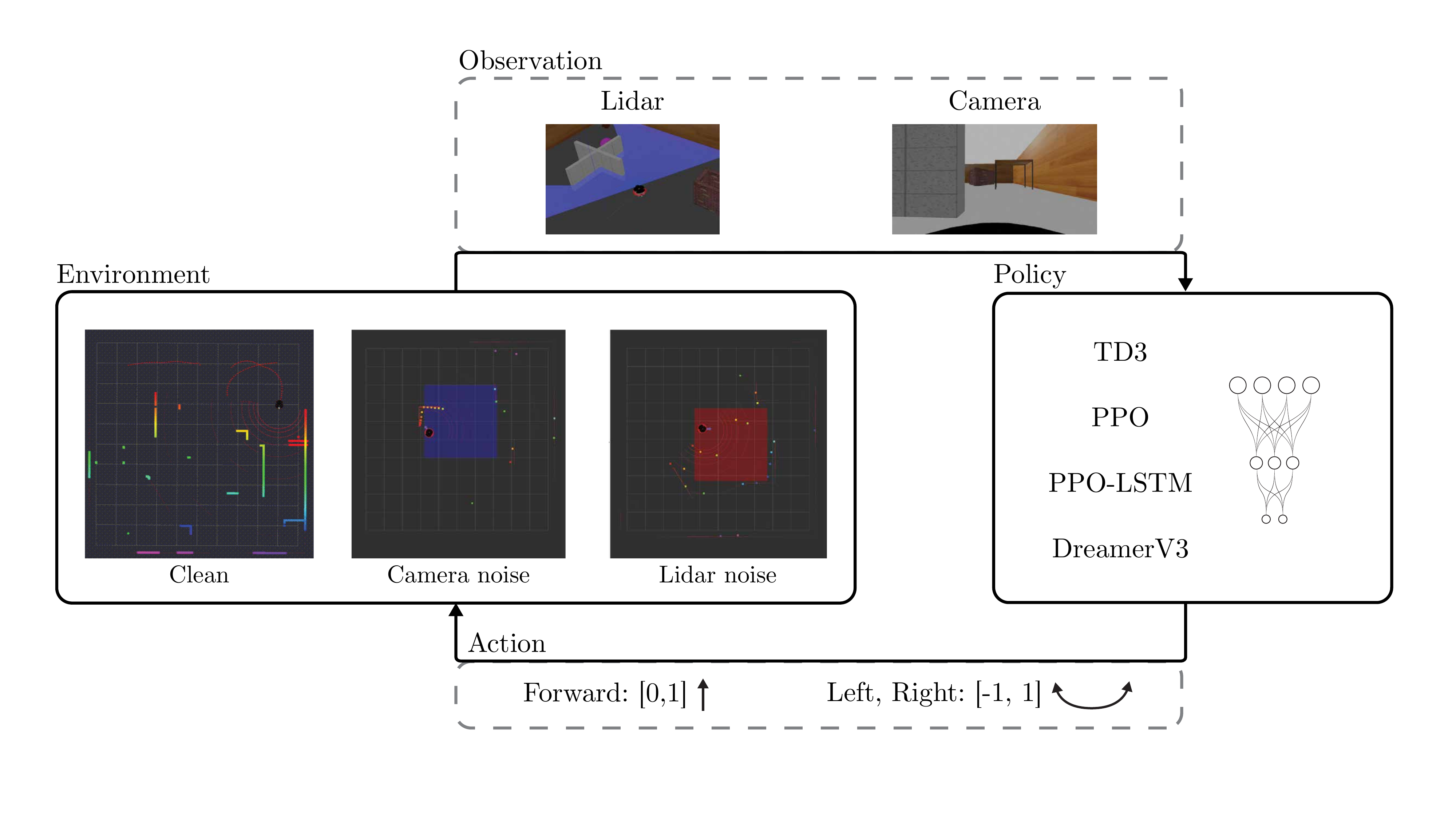}
\caption{Experimental diagram showing the process of the presented experiments. The environment consists of clean, camera noise, and Lidar noise variants. The environment outputs the observation in form of two modalities, Lidar and camera. Multiple algorithms are then trained to navigate to the goal whilst avoiding the obstacles.}
\label{fig:experimental_diagram}
\end{figure}

As the focus of this study is on studying adversarial sensor perturbations and testing different sensor suites, the changes to the environment are described in section \ref{subsection:changes_to_env}. The benchmarking procedure, of training the different RL algorithms - TD3, PPO, PPO LSTM, and DreamerV3, using different modalities and with differing sensor noise, is illustrated in figure \ref{fig:experimental_diagram}.
The reinforcement learning algorithms used in the benchmark are described in section \ref{subsection:RL}. Training experimental settings, including hyperparameters, are shown in section \ref{subsection:sim_params}. The evaluation process is described in section \ref{subsection:evaluation}.

\subsection{Changes to the DRL-Robot-Navigation Environment}\label{subsection:changes_to_env}
To enable our experiments, the \textit{DRL-Robot-Navigation} environment required several changes. 
A primary change to the environment is the addition of a vision-based observation space. To do this, the image from the robot is outputted from the environment, and a visual marker is added to mark the position of the goal. A pink sphere of diameter 0.5 m is added where the goal resides and is shown in Fig. \ref{fig:methodology:drl_env} (top left image).  

The collision system is updated, as in the original implementation, it relied on the reading from the Lidar sensor. This is not a reliable way of detecting collisions, and adding noise to the Lidar sensor could trigger collisions. Instead, we generate a flag and detect any physical collisions between objects by triggering ROS topic that can be subscribed. 

We added areas that contain sensor noise: separately for Lidar and camera. The areas are fixed-size rectangles that spawn at a random point of the map at the start of each episode. The way the sensor faults are modelled is described further in section \ref{subsection:sensor_modelling}.

We updated the reward system. Originally, the agent received 100 reward for reaching the goal and -100 for a collision with a wall. Otherwise, a function of the actions (penalizing the robot for inaction) and the minimum distance from a wall (based on the Lidar observation, if the robot is closer than 1 m to a wall) was returned. 
First, we normalized the reward between +1 and -1. Next, we added a penalty for taking a step: of size -1 divided by the maximum number of steps. This was done to prevent a local minima where the robot doesn't move and is in line with other gymnasium environments \cite{towers_gymnasium_2024}. 
We also considered the function of penalizing inaction and Lidar distance to the wall to be unnecessary. Instead, we consider that these items - inaction and obstacle avoidance - should be learned on the policy side from observation without reward shaping.

\subsubsection{Modelling Adversarial Sensor Perturbations}\label{subsection:sensor_modelling}

\begin{figure}[h]
    \subfigure[\label{fig:methodology:Lidar_noise}]{\includegraphics[width=.5\textwidth]{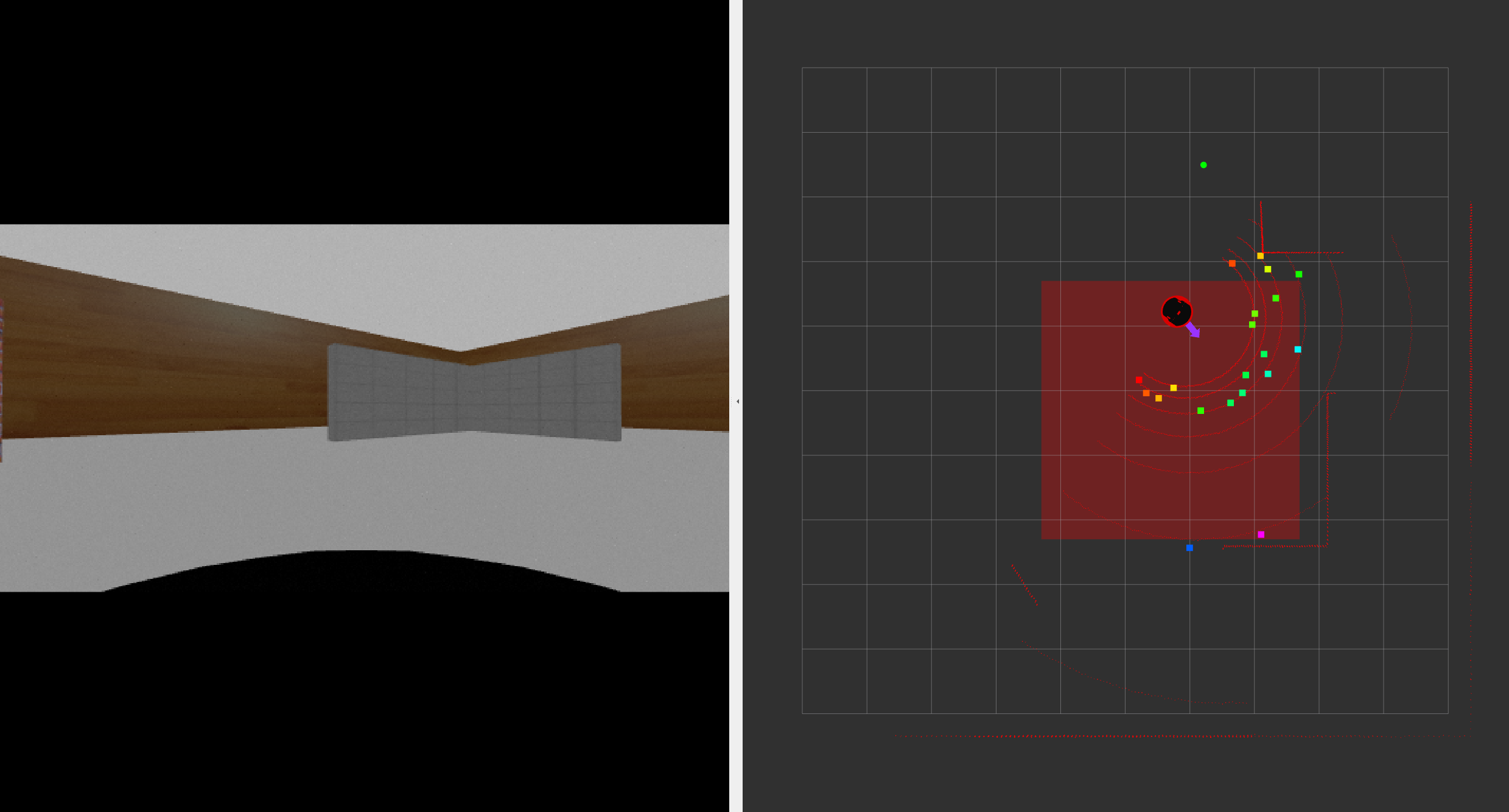}}%
    \quad  
    \subfigure[\label{fig:methodology:camera_noise}]{\includegraphics[width=.5\textwidth]{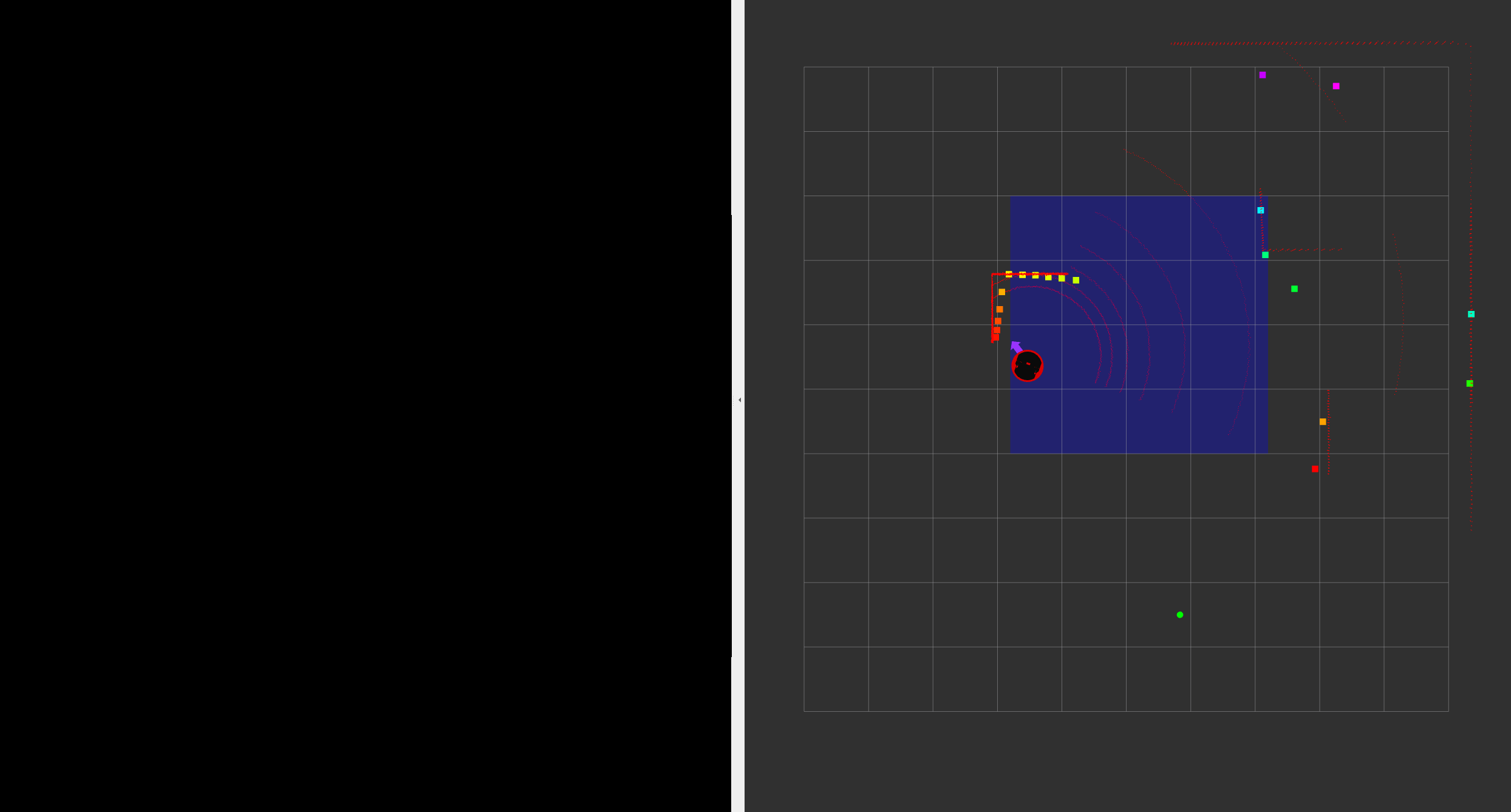}}%
    \quad  

     \caption{Example of a robot in the Lidar noise area, visualized by RViz. Once inside the red Lidar noise area, the discretized Lidar points have Gaussian noise added to them and no longer provide accurate representations of the environment.  
     Example of a robot in the camera noise area, visualized by RViz. Once inside the blue camera noise area, the image feed (shown on the left) has all pixels turned to black - this simulates a camera failure. 
     }
    \label{fig:methodology:noise_examples}
\end{figure}

Autonomous vehicles often use a camera as a primary sensor for navigation, but physical sensor attacks are rarely considered. 
Kim et al. \cite{kim_review_2022} perform a review of drone attacks, including laser attacks on the camera sensor. They find that when applied, the camera pixels are distorted. This shows that camera sensors are susceptible to physical attacks. 
Camera attacks are modelled as a complete failure, i.e. all the pixels are turned black (0) whilst inside of the camera noise rectangle. Whilst this is not explicitly modelled after a physical attack, it is a possible temporary failure mode for a camera and is shown in Fig. \ref{fig:methodology:camera_noise}.
Lidar attacks are modelled by adding a strong Gaussian noise, ranging between 0-5 meters to each of the Lidar points. 
Sensor noise is applied evenly throughout the noisy area. A Lidar sensor perturbation is shown in Fig. \ref{fig:methodology:Lidar_noise}

\subsection{Reinforcement Learning Benchmark}\label{subsection:RL}

A benchmark of the following RL methods is performed on the \textit{DRL-Robot-Navigation} environment:
\begin{itemize}
    \item \textbf{Twin Delayed DDPG (TD3)} \cite{fujimoto_addressing_2018} is an off-policy, model-free algorithm designed to work on continuous action spaces. It employs two Q-value approximators (critics) and uses the smaller of the two Q-values to update the policy, which helps to mitigate the problem of overestimation bias. TD3 also introduces a delay in updating the policy network to reduce the variance of policy updates and adds noise to the target action to encourage exploration. The hyperparameters used to train TD3 are shown in table \ref{table:TD3_params}. 
    \item \textbf{Proximal Policy Optimization (PPO)} \cite{schulman_proximal_2017} is an on-policy, model-free algorithm designed to work on continuous spaces. \textbf{Recurrent PPO}, a modification of PPO with the introduction of a long short-term memory (LSTM) layer \cite{hochreiter_long_1997}, is also used. The hyperparameters used to train PPO and Recurrent PPO are shown in table \ref{table:PPO_params}. The recurrent architecture is similar to the PPO, but contains an LSTM layer before the Actor-Critic Multilayer Perceptron (but after the convolutional layers for the camera observation).
    \item \textbf{DreamerV3} \cite{hafner_mastering_2023} is an off-policy, model-based algorithm that learns a world model from the experience by incorporating an autoencoder that encodes the input into a discrete representation. It uses recurrence and predicts the dynamics, reward, and the continuity of the episode. The hyperparameters are fixed, hence tuning is not required. It has shown to outperform PPO on several environments, including navigation on DeepMind Lab \cite{beattie_deepmind_2016}. Based on this, it is expected that DreamerV3 should outperform the other algorithms on our navigation problems. The default model is used. 
\end{itemize}

\begin{figure}[!t]
\centering
\includegraphics[width=\textwidth]{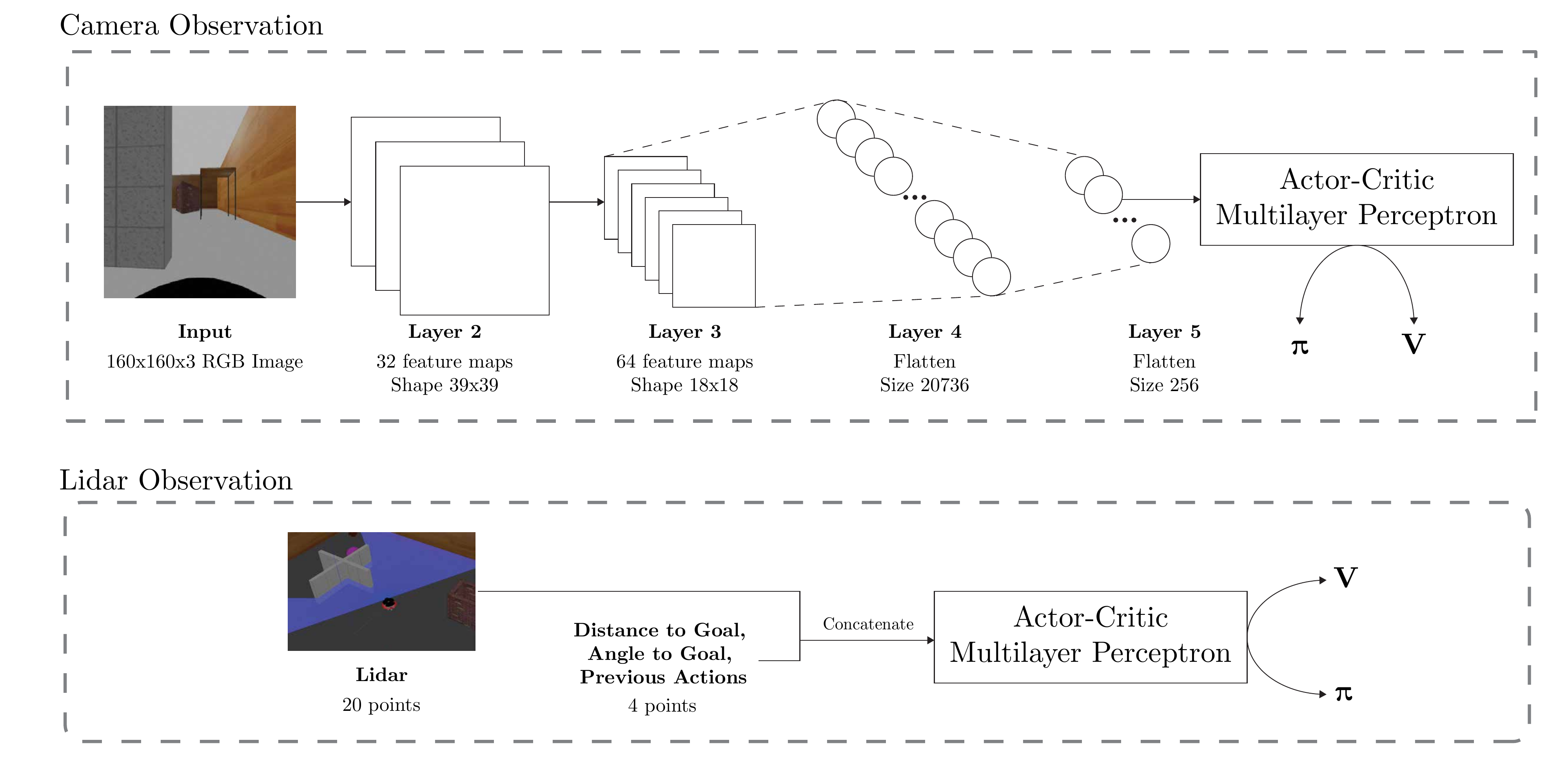}\caption{Camera and Lidar observation networks. In camera observation models (TD3, PPO, PPO-LSTM) the image is first passed through convolutional layers and flattened to 256 features to reduce the dimensionality of the input. These features are then input to the actor-critic MLP. For the Lidar observation models (TD3, PPO), the input is already lowly dimensional and so the 20 Lidar readings are concatenated with the distance to goal, angle to goal, and previous actions and input to the MLP. 
}
\label{fig:methodology:RL_network}
\end{figure}

Stable-Baselines3 \cite{raffin_stable-baselines3_2021} implementations of TD3, PPO, and recurrent PPO were used for the experiments. The general setup for TD3 and PPO is shown in Fig. \ref{fig:methodology:RL_network}, showing the networks for both camera and Lidar observations. For the camera observation, the image is initially passed through a convolutional network to reduce the dimensionality to 256 features. Those features are then passed into the actor-critic MLP. In the case of the recurrent PPO, the features are first passed through an LSTM layer, before going to the actor-critic. For the Lidar, because it only contains 24 data points, the features are passed straight to the actor-critic.
We perform: 
\begin{itemize}
    \item A benchmark of the RL methods.
    \item A learning-rate discovery to find the optimal LR for each algorithm as it is the most sensitive hyperparameter (apart from Dreamerv3 as it does not require hyperparameter tuning). This is shown in the appendix. 
    \item A study into the performance of the agents across different observation spaces in sensor-perturbed areas.
\end{itemize}

\begin{table}[!t]
\centering
\caption{TD3 training parameters (final values are bolded)}
\label{table:TD3_params}
\begin{tabular}{|l|l|}
\hline
{Parameter} & {Value} \\[1mm]
\hline
Learning Rate & 0.03 / 0.003 / \textbf{0.0003} / 0.00003 (see LR study) \\
\hline
Tau & 0.005 \\
\hline
Learning Starts (after, steps) & 5000 \\
\hline
Batch Size & 16384 (Lidar), 256 (Camera) \\
\hline
Discount Factor & 0.99  \\
\hline 
Exploration Noise & 0.3 \\
\hline
\end{tabular}
\end{table}

\begin{table}[!t]
\centering
\caption{PPO and PPO LSTM training parameters (final values are bolded)}
\label{table:PPO_params}
\begin{tabular}{|l|l|}
\hline
{Parameter} & {Value} \\[1mm]
\hline
Learning Rate (Lidar) & 0.03 / \textbf{0.003} / 0.0003 / 0.00003 (see LR study) \\
\hline
Learning Rate (Camera) & 0.03 / 0.003 / \textbf{0.0003} / 0.00003 (see LR study) \\
\hline
Batch Size & 256  \\
\hline
Discount Factor & 0.99  \\
\hline 
\end{tabular}
\end{table}

\subsection{Experimental Scenario Settings}\label{subsection:sim_params}
The experimental settings are shown in table \ref{table:sim_params}. The Lidar observation space comprises 20 discrete points, taken at even intervals in a 180° arc around the robot. The camera image produces a 160x160 px RGB image and 64x64 px RGB image for DreamerV3 (the input to the DreamerV3 is required to be a power of two\footnote{\url{https://github.com/danijar/dreamerv3/issues/12}}). Lidar noise is Gaussian, and camera noise results in a complete blackout. The agent position input (for Lidar modality only) is fully accurate. The goal position for Lidar is random, and for the camera it is static. The time delta between steps is 0.05 seconds - this is the delay between which the agent makes actions. Note that the simulation runs at 4.0 times real-time speed, which results in this time step being 0.2 seconds, i.e. the agent runs at effectively 5 Hz in real time. The reward function produces a sparse reward when the agent reaches the goal, a penalty for colliding with a wall, and a small penalty each step to prevent inaction. The maximum steps per episode is 50. 

There exist several random parameters within the simulation that affect the learning process. Randomness plays a key role, as shown by Mirowski et al. \cite{mirowski_learning_2017}. In their simulation, the RL agents outperformed humans on static small mazes but struggled on large complex mazes (with random parameters such as random agent spawning). Another difference between \textit{DRL-Robot-Navigation} and DeepMind Lab is that our environment penalizes collisions with the wall, preventing wall-following strategies. 

Camera policies require longer training compared with Lidar (greater than 500k steps) which exceeds the computational budget for this study. To simplify it, the goal is spawned at the centre of the map, instead of being spawned randomly. The robot is still spawned with a random position and orientation. This is akin to the 'static maze' from DeepMind Lab, as tested by Mirowski et al. Their 'random goal maze' is akin to our random goal position that we test during our Lidar experiments (albeit DeepMind Lab did not penalize wall collisions). 

Generally, apart from DreamerV3, the camera policies likely take longer to converge to a solution due to increased observation space: even after convolutional layers, the state contains 256 features compared with the much simpler 24 points for Lidar (including previous actions, distance to goal, and angle to goal). 

The original implementation of \textit{DRL-Robot-Navigation} environment randomizes the starting position and orientation of the robot, as well as the position of the goal at the start of each episode. 

To improve the convergence of camera policies, in order to better understand the sensor denial effects, we simplify the environment for our camera experiments. We fix the position of the goal to always be in the centre of the maze. This allows for less stochasticity in the environment and enables for quick convergence of the policy. 
For the Lidar experiments, we keep the spawn position of the goal and the robot random.

\begin{table}[!t]
\centering
\caption{Simulation Parameters}
\label{table:sim_params}
\begin{tabular}{|p{3cm}|p{5cm}|} 
\hline
\textit{} & \textit{\textbf{Parameter details}} \\ \hline
Lidar Measurements & \begin{tabular}[c]{@{}l@{}}180° LiDAR data discretized \\ into 20 intervals \end{tabular} \\ \hline
Camera Measurements & \begin{tabular}[c]{@{}l@{}}150px x 150px RGB image \\ 64px x 64px RGB image (Dreamer) \end{tabular}\\ \hline
Attack Type Lidar & Gaussian noise added to each discrete point \\ \hline
Attack Type Camera & Complete failure \\ \hline
Agent Position & \begin{tabular}[c]{@{}l@{}}Fully accurate\\ (ground truth from Simulation)\end{tabular} \\ \hline
Goal Position & \begin{tabular}[c]{@{}l@{}}Lidar: random \\ Camera: static \end{tabular}\\ \hline
Delay between steps & 0.05 seconds \\ \hline
Reward function & \begin{equation}
R(s, a) =
\begin{cases}
1, & \text{goal reached} \\
-1, & \text{collision} \\
-1/50, & \text{otherwise}
\end{cases}
\end{equation} \\ \hline
Max. Episode Steps & 50 \\ \hline
\end{tabular}
\end{table}

\subsection{Training and Evaluation}\label{subsection:evaluation}

The reinforcement learning (RL) models described in section \ref{subsection:RL} are trained with the following configurations and environments:
\begin{itemize}
    \item Lidar (this includes distance and angle to goal). The models are trained on variety of 0x0 (no noise) maps, as well as maps with varying amounts of noise: 3x3, 5x5, and 7x7 metre noise zone.
    \item Camera only, static goal. The models are trained on variety of 0x0 (no noise) maps, as well as maps with varying amounts of noise: 3x3, 5x5, and 7x7 metre noise zone.
\end{itemize}
Unless otherwise stated, camera configurations contain a static goal to enable better convergence, as described in section \ref{subsection:sim_params}. 
Each model is then evaluated on maps specified in the section \ref{subsection:evaluation}, but generally this consists of being tested on each of the environments of differing noise size: 0x0, 3x3, 5x5, 7x7. 
In some cases, the model is evaluated on a simplified version of the map, as described in section \ref{subsection:evaluation}.

\begin{figure}[!t]
\centering
\includegraphics[width=\textwidth]{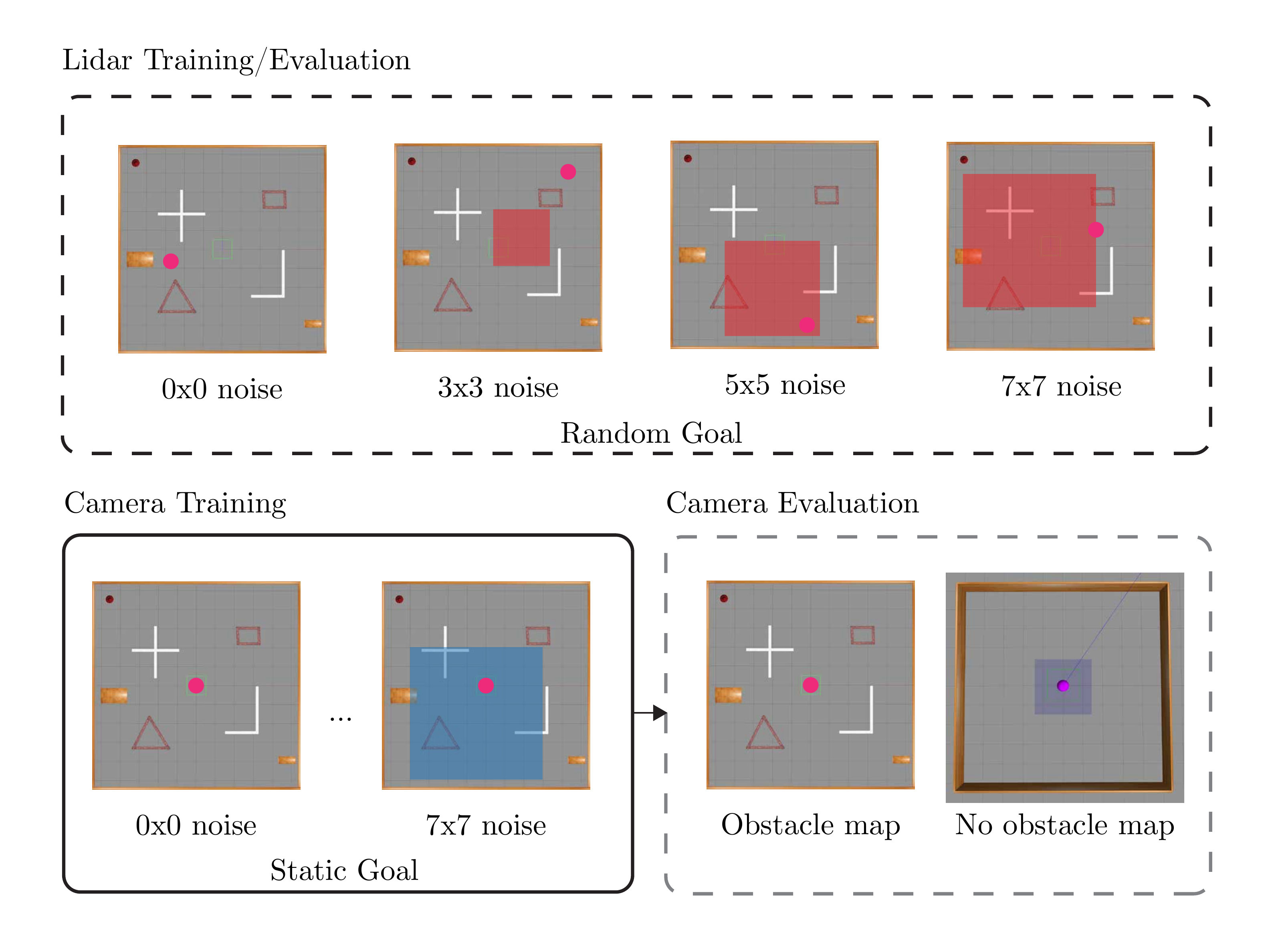}\caption{
Training and evaluation scenarios for Lidar and camera sensors. For Lidar scenarios, the models are trained on a variety of 0x0 (vanilla) maps and maps with varying amounts of sensor perturbation: 3x3, 5x5, and 7x7 metre sensor noise zones. These zones contain Gaussian noise which is added to the sensor readings. These trained policies are then evaluated on the same map.
Likewise, for camera scenarios the models are trained on a variety of 0x0 (vanilla) maps and maps with varying amounts of sensor perturbation: 3x3, 5x5, and 7x7 metre sensor denial zones. Inside these zones, the camera sensor completely fails (all pixels turn to black). The models are trained on the default map, with a static goal in the middle, and the blue areas in which the camera completely fails. These policies are then evaluated on the default map with varying degree of noise. A second evaluation is performed on a map with no obstacles, to better understand the learned policies in sensor denied areas.
}
\label{fig:methodology:training_evaluation_setup}
\end{figure}

Fig. \ref{fig:methodology:training_evaluation_setup} shows the outline of the training and evaluation setup for Lidar and camera environments.

\subsubsection{Camera Evaluation Maps}\label{subsection:evaluation_maps}

\begin{figure}[h]
    \subfigure[\label{fig:methodology:evaluation_map_noise_3}]{\includegraphics[width=.3\textwidth]{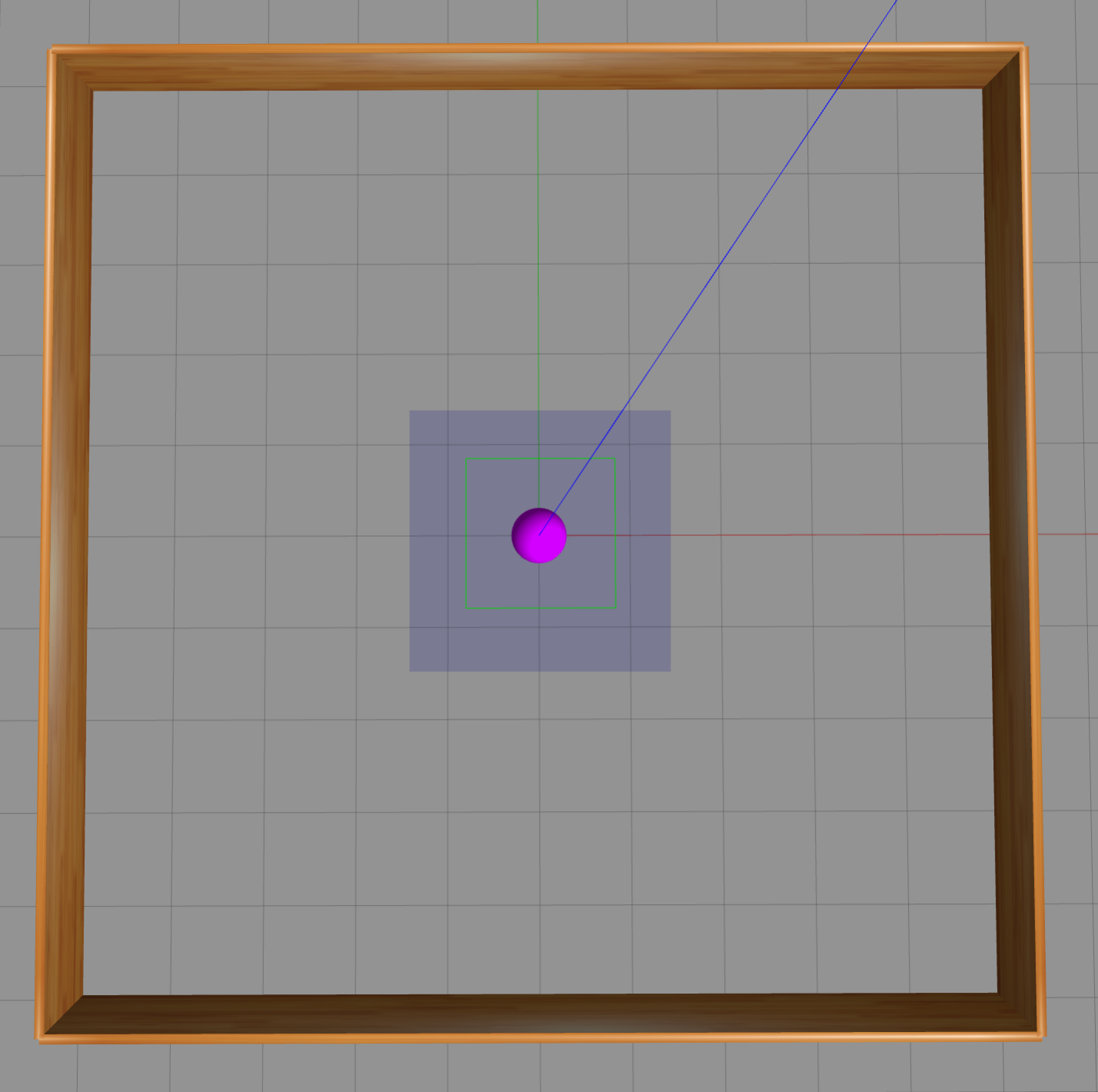}}%
    \quad  
    \subfigure[\label{fig:methodology:evaluation_map_noise_5}]{\includegraphics[width=.3\textwidth]{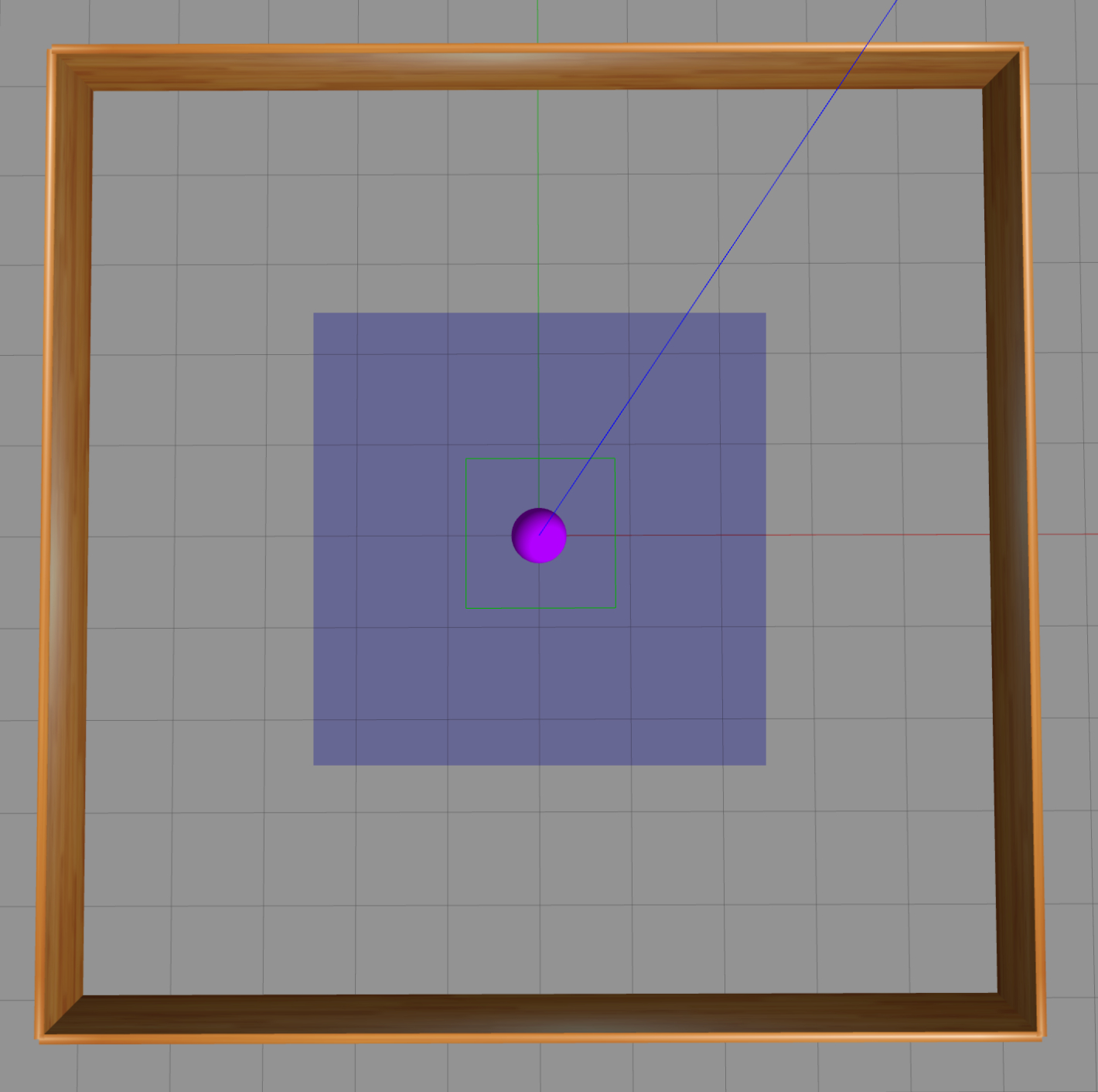}}%
    \quad  
    \subfigure[\label{fig:methodology:evaluation_map_noise_7}]{\includegraphics[width=.3\textwidth]{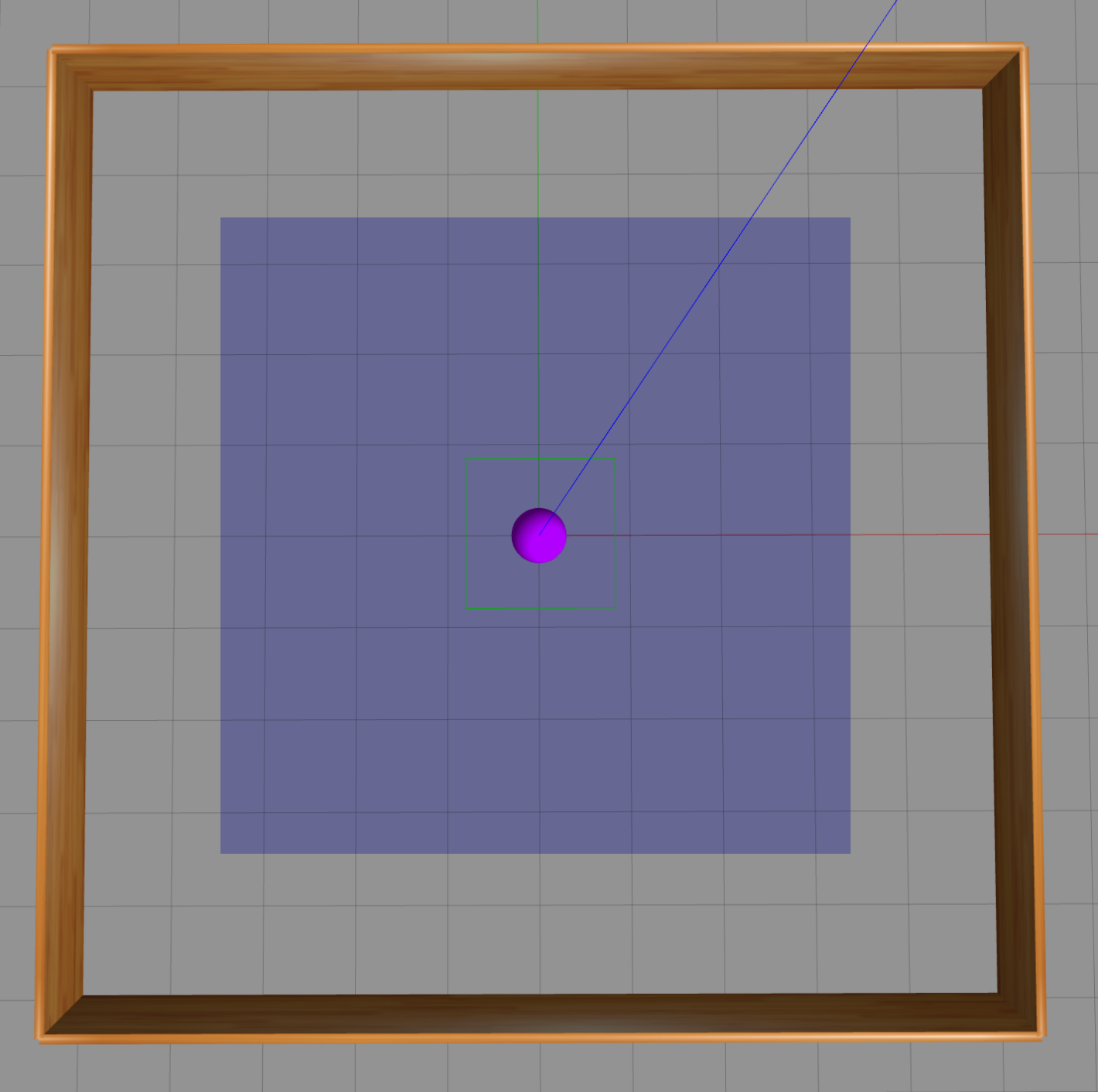}}%
     \caption{Simplified version of the \textit{DRL-Robot-Navigation} map with no obstacles, with a static goal in the centre and a sensor perturbation area surrounding it. \textbf{(a)} shows the 3x3 sensor denied area, \textbf{(b)} shows the 5x5 sensor denied area, and \textbf{(c)} the 7x7 sensor denied area. The robot is spawned at a random position and orientation around the map. This map is only used to evaluate the camera policies, and not Lidar policies. }
    \label{fig:methodology:evaluation_simplified_map}
\end{figure}

For evaluation of camera policies, two different maps are used:
\begin{itemize}
    \item \textit{DRL-Robot-Navigation} map (same as training). 
    \item A simplified \textit{DRL-Robot-Navigation} map no obstacles, shown in Fig. \ref{fig:methodology:evaluation_simplified_map}. The goal is always static in the centre. The noise area is static in the centre of the map. 
\end{itemize}

The purpose of using the simplified map is to better understand the behaviour of the policies. As the obstacles are taken out, the only task left is to get through the sensor denied area to the goal. As the noise is static, it always obstructs the goal. In this evaluation task we are ignoring the ability of the agent to avoid obstacles, but are focusing on whether the agent has learned to get through the noise area.
The simplified map is only used for evaluation and is never used for training of the policies. By default, the \textit{DRL-Robot-Navigation} is used for evaluation. The use of the simplified maps is explicitly specified in the results section.

\section{Results}\label{section:Results}
Several experiments are conducted to benchmark the different DRL algorithms on the environment and understand how different training regimes (vanilla vs adversarial) affect the performance of the models in environments with varying environmental perturbation size. While initially training of the Lidar and camera modalities is compared in section \ref{section:results:modality_study}, the rest of the results section is split into two parts, based on modalities. The first part, in section \ref{section:results:lidar_observation_quantitative} follows how PPO and TD3 perform on in terms of quantitative results, and then qualitatively in section \ref{section:results:lidar_observation_qualitative} by understanding the paths followed by the policies, and their differences under different training regimes. 

The camera observation section begins with a direct training comparison of the different algorithms on the vanilla environment in section \ref{section:results:algorithm_training_comparison}, continues with a quantitative benchmark of the different models with sensor denial in section \ref{section:results:camera_observation_regimes}, and the models are evaluated on the no obstacle map in section \ref{section:results:navigation_noise_camera_no_obstacle}. Finally, the results are discussed with respect to literature in section \ref{section:results:discussion}.

\subsection{Modality Study}\label{section:results:modality_study}
\begin{figure}[h]
\centering
\includegraphics[width=\textwidth]{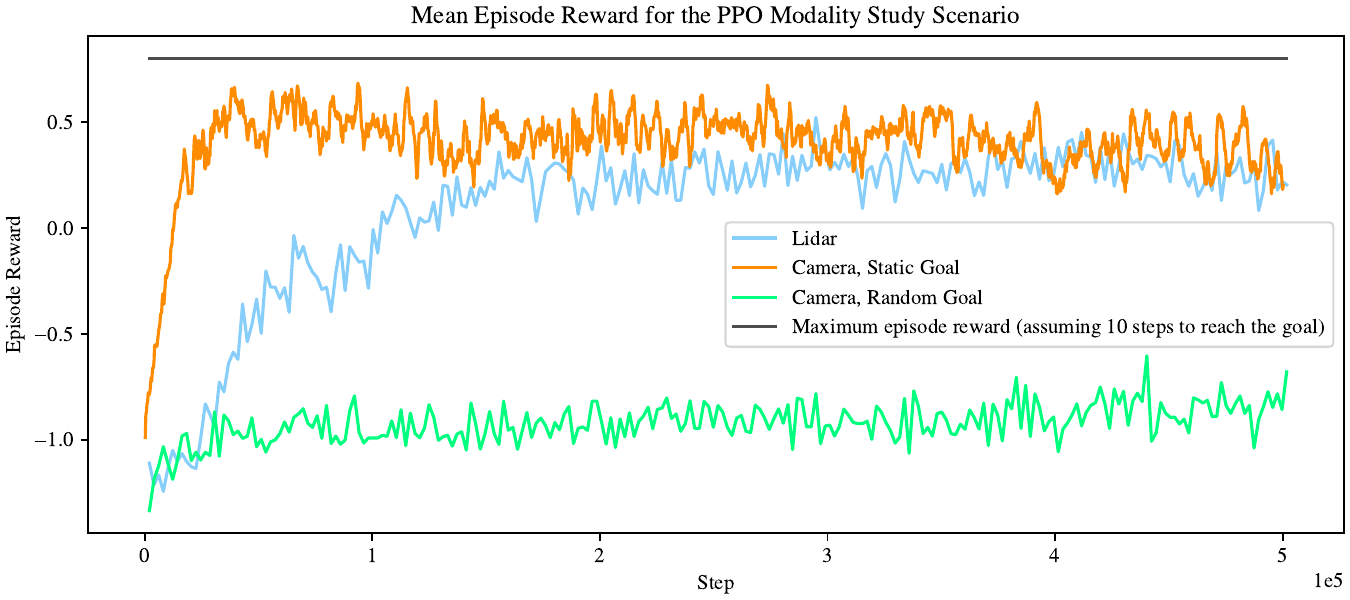}
\caption{Modality study using PPO for Lidar and camera. The camera was trained on static and random goals. Each of the models was trained for 500,000 steps.}
\label{fig:results:modality_study}
\end{figure}

The modality study shown in Fig. \ref{fig:results:modality_study} shows the differences in model performance for the PPO algorithm across Lidar (with a random goal), camera with a static goal, and camera with a random goal. Lidar and camera with a static goal are quick to converge to a solution close to the maximum episode reward. Camera with a random goal does not converge to a solution in the given amount of steps. As we attempt to understand the effects of sensor perturbations, we will continue using camera with a static goal for the camera noise experiments.

\subsection{Lidar Navigation with Noise}\label{section:results:lidar_observation_quantitative}

Although different algorithms are proposed in section \ref{subsection:RL}, we were unable to find an appropriate learning rate to get PPO-LSTM to converge as shown in appendix \ref{fig:appendix:ppo_lstm_lr}. While it is possible to train DreamerV3 on the Lidar observation space, it is a large model for such a small observation space and we did not find that its performance increased over the other models. Hence, only TD3 and PPO are compared in this section. 

\begin{figure}[h]
\centering
\includegraphics[width=\textwidth]{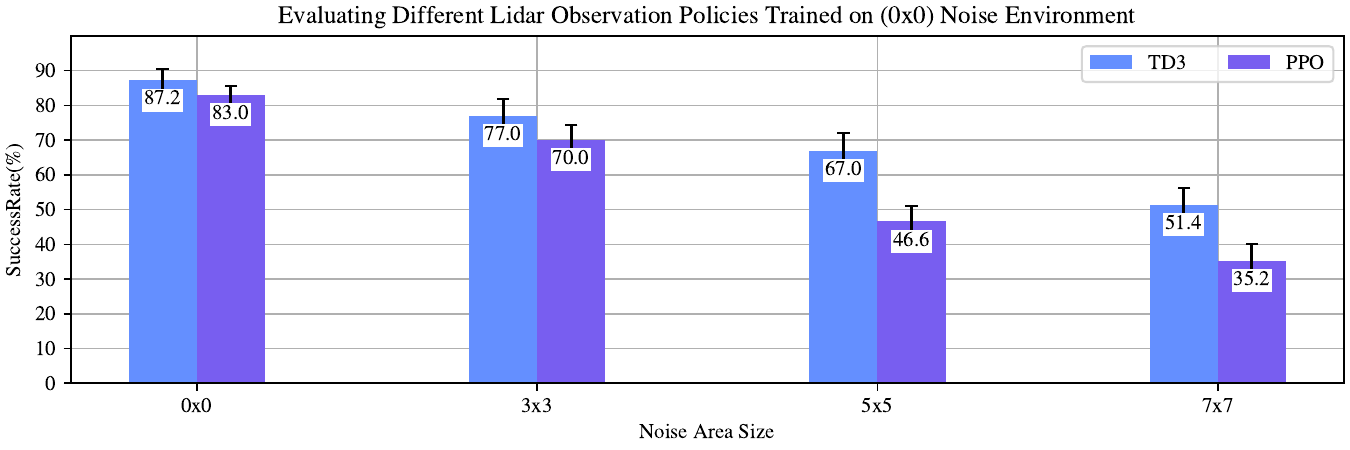}
\caption{Lidar noise area study for algorithms trained on the default environment with no noise. The algorithms were then evaluated on the default environment (with a random goal) with changing noise area sizes, as shown on the x-axis. The evaluation was performed for 100 episodes and repeated 5 times to establish a standard deviation.}
\label{fig:results:lidar_noise_0x0}
\end{figure}

Fig. \ref{fig:results:lidar_noise_0x0} shows the evaluation results of the TD3, and PPO using the Lidar observation space on the default map. The algorithms are trained on the 0x0 (vanilla) map, and evaluated on environments of increasing Lidar noise area: 0x0 (vanilla), 3x3, 5x5, and 7x7 - referring to the size of the rectangle of the noise area in meters. 
TD3 outperforms PPO across the different evaluations. It achieves a mean success rate of 87.2\% on the 0x0 evaluation scenario, compared with 83.0\% for PPO. The performance differential between the two models generally increases with larger noise area sizes. When evaluated on the 7x7 noise area size, TD3 achieves 51.4\%, while PPO achieves 35.2\%.

\begin{figure}[h]
\centering
\includegraphics[width=\textwidth]{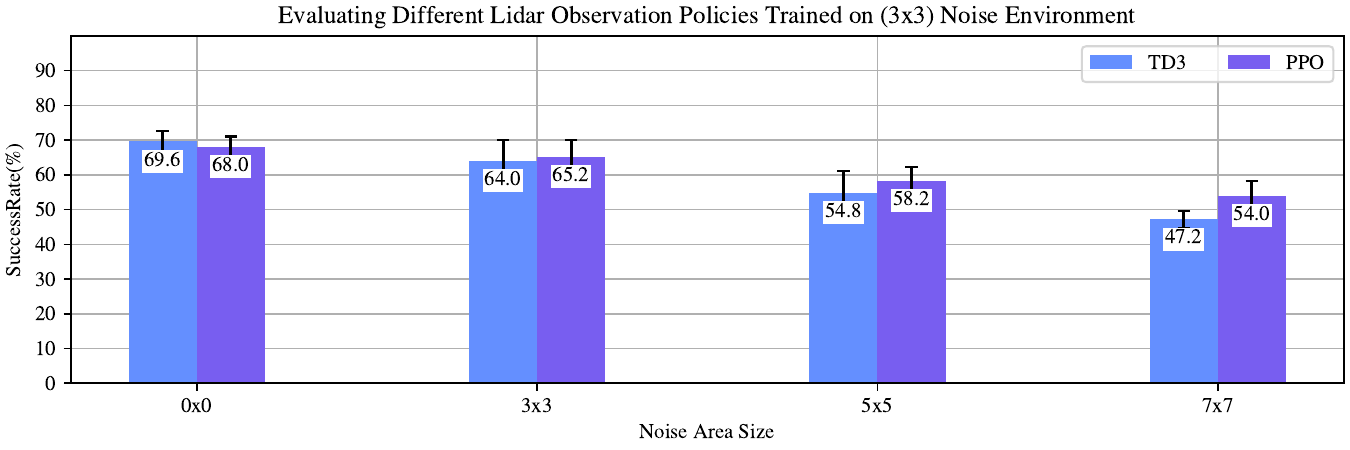}
\caption{Lidar noise area study for algorithms trained on the default environment with random 3x3 noise areas and random goal. The algorithms were then evaluated on the same map with different noise area sizes, as shown on the x-axis. The evaluation was performed for 100 episodes and repeated 5 times to establish a standard deviation.}
\label{fig:results:lidar_noise_3x3}
\end{figure}

Fig. \ref{fig:results:lidar_noise_3x3} shows the same evaluation results of the TD3, and PPO, with the algorithms trained on the 3x3 noise map. In these results, the PPO closely matches the performance of TD3 across the evaluations, outperforming it on larger noise area sizes. 
When evaluated on the vanilla 0x0 scenario, TD3 achieves a mean success rate of 69.6\% and PPO achieves 68.0\%. This is a performance drop from the models trained on the vanilla 0x0 environments. 
When evaluated on the 7x7 scenario, the PPO model achieves a mean success rate of 54.0\%. This is an improvement from the PPO model trained on the vanilla 0x0 environment and slightly outperforms the TD3 model trained on the same regime. 
The TD3 model trained on the 3x3 noise environment shows performance drops across all evaluation scenarios compared with the TD3 model trained on the 0x0 environment.

\subsection{Lidar Noise Training Regimes, Qualitative Results}\label{section:results:lidar_observation_qualitative}
As shown in section \ref{section:results:lidar_observation_quantitative}, the training regimes affect the evaluation performance of the algorithms. To better understand the behaviour of Lidar-specific policies, we can more closely investigate the effects of training in noisy regimes for each algorithm and investigate the trajectories of the robot during evaluation. 

\begin{figure}[h]
\centering
\includegraphics[width=\textwidth]{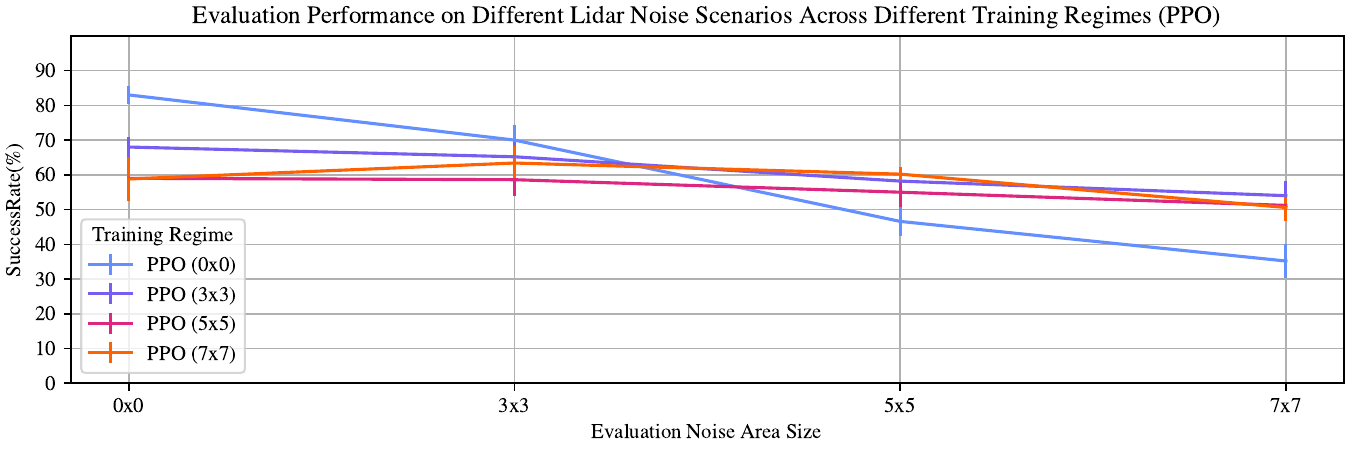}
\caption{PPO policies trained on different Lidar noise regimes evaluated on the default map (with random goal) with varying noise area sizes. The policies were evaluated for 100 steps 5 times, and the plot shows a mean and the standard error across those 5 repeated runs. }
\label{fig:results:lidar_regimes_ppo_policies}
\end{figure}

Fig. \ref{fig:results:lidar_regimes_ppo_policies} shows the PPO policies trained on different Lidar noise regimes evaluated on the default map. 
It shows that the vanilla PPO (0x0) performs the best when evaluated on a map with no noise, with a mean success rate of 83\%. The policies trained on noisy environments (PPO 3x3, 5x5, and 7x7) begin to outperform the vanilla policy when evaluated on the 5x5 noise map. While the noisy policies do not significantly degrade in performance as more noise is added to the environment, they show a consistent drop in performance compared with the vanilla model when evaluated on the 0x0 no noise environment.
This is likely due to the noisy policies following a riskier strategy, sacrificing collision avoidance capabilities to improve the navigation performance. To better understand this, we can qualitatively investigate the evaluation runs by plotting their paths, crashes and successful navigation.

\begin{figure}[h]
    \subfigure[\label{fig:results:paths_ppo_0x0_eval_0x0_lidar}]{\includegraphics[width=.49\textwidth]{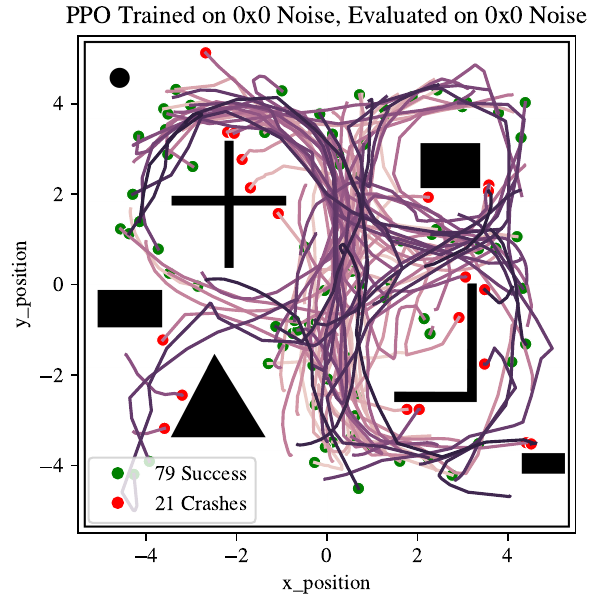}}%
    \quad  
    \subfigure[\label{fig:results:paths_ppo_0x0_eval_7x7_lidar}]{\includegraphics[width=.49\textwidth]{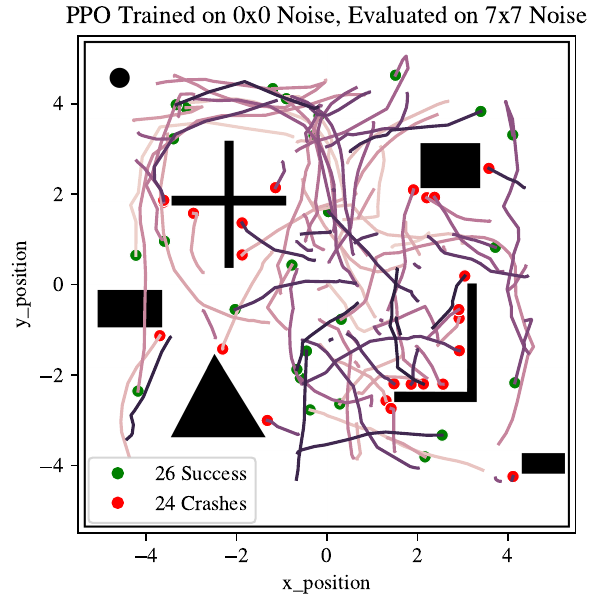}}%
    \quad  
    \subfigure[\label{fig:results:paths_ppo_3x3_eval_0x0_lidar}]{\includegraphics[width=.49\textwidth]{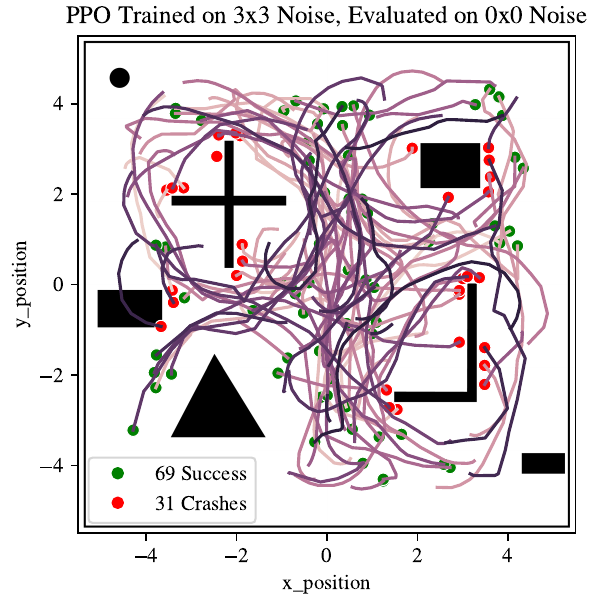}}%
    \quad  
    \subfigure[\label{fig:results:paths_ppo_3x3_eval_7x7_lidar}]{\includegraphics[width=.49\textwidth]{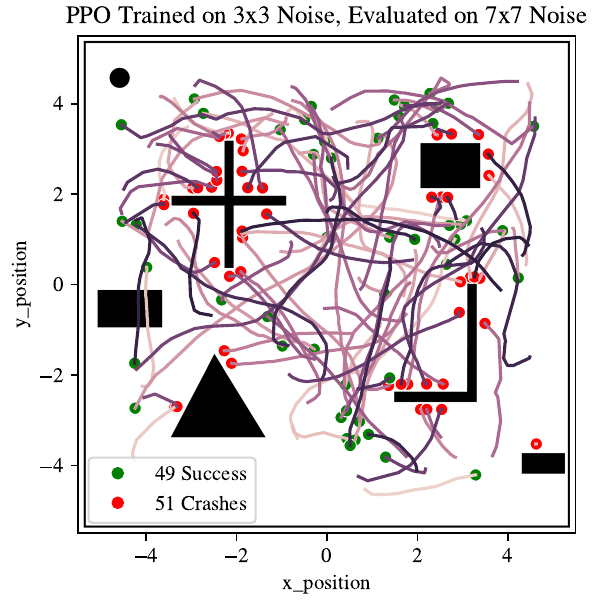}}%
     \caption{100 path plots generated during evaluation by vanilla policy (PPO 0x0) evaluated on 0x0 noise map \textbf{(a)}, vanilla policy (PPO 0x0) evaluated on 7x7 noise map \textbf{(b)}, noise policy (PPO 3x3) evaluated on 0x0 noise map \textbf{(c)}, and noise policy (PPO 3x3) evaluated on 7x7 noise map \textbf{(d)}. The red dot shows a collision and the green dot shows a successful navigation to the goal.}
    \label{fig:results:qualitative_lidar_ppo_regime_runs}
\end{figure}

Fig. \ref{fig:results:qualitative_lidar_ppo_regime_runs} shows the 100 path plots generated during the evaluation of 2 different policies (PPO 0x0 and PPO 3x3) on two different evaluation maps (0x0 and 7x7). Fig. \ref{fig:results:paths_ppo_0x0_eval_0x0_lidar} shows the vanilla PPO evaluated on the no noise map. The same policy is evaluated on a map with random 7x7 noise areas in Fig. \ref{fig:results:paths_ppo_0x0_eval_7x7_lidar}. The policy is unable to reach the targets and shows a significant decrease in performance (from 79 successes to 26 successes). It also shows that in half of the runs, the policy reached the maximum episode time step limit and it neither crashed nor reached the target. Compared with the evaluation on the no noise map, the paths tend to be much shorter, showing that the policy prefers to follow a 'safe' strategy of not doing anything when inside the noisy areas. 
In contrast, when trained on a little noise (3x3 random areas) and evaluated on no noise as shown in Fig. \ref{fig:results:paths_ppo_3x3_eval_0x0_lidar}, the overall performance drops from 79 successful episodes to 69, and the number of crashes increases from 21 to 31. But, when evaluated on the 7x7 noise area as shown in Fig. \ref{fig:results:paths_ppo_3x3_eval_7x7_lidar}, the policy no longer follows a 'safe' strategy and instead opts for a 'riskier' approach. This increases the number of successful episodes from 26 to 49 but also increases the amount of crashes from 24 to 51.

Overall, the safe policy did not result in better overall performance because the reward for reaching the max number of steps is -1, like with crashing, it is -1. But, because the risky policy results in more successful runs resulting in the positive +1 reward, it is the more optimal policy for the noisy evaluation maps.
It is an interesting observation that the riskier strategies consistently emerged across the PPO runs when trained on noisy maps, while when not trained on any noise, the policy followed a safer strategy of letting the episodes run out when it was not certain of its sensor readings, as shown in Fig. \ref{fig:results:lidar_regimes_ppo_policies}. This is likely due to the noisy sensor readings falling outside of the previously observed domain for the policy trained on the non noisy environment.

\begin{figure}[h]
\centering
\includegraphics[width=\textwidth]{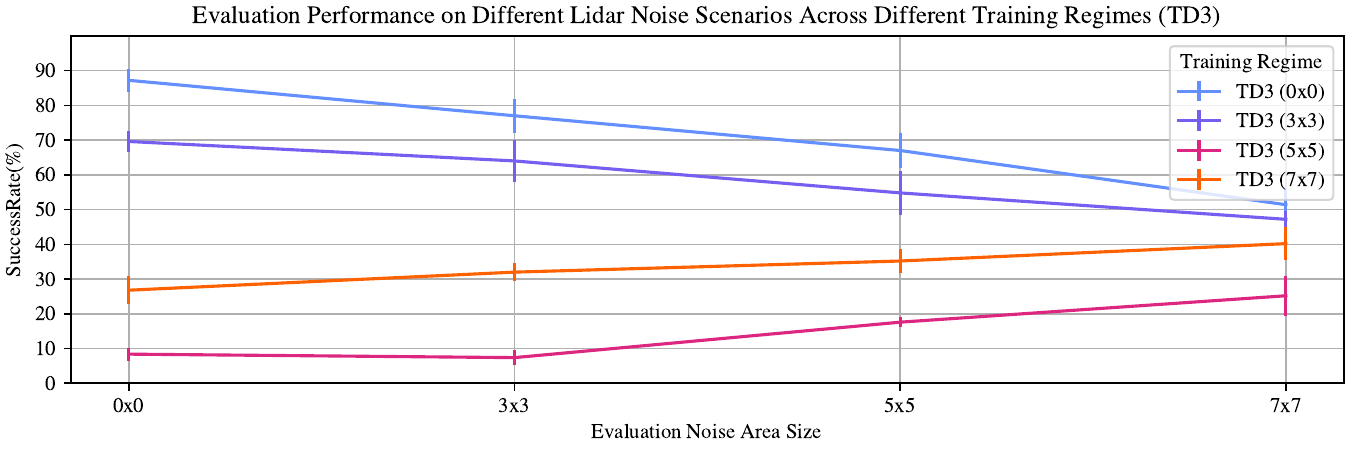}
\caption{TD3 policies trained on different Lidar noise regimes evaluated on the default map (with random goal) with varying noise area sizes. The policies were evaluated for 100 steps 5 times, and the plot shows a mean and the standard error across those 5 repeated runs.}
\label{fig:results:lidar_regimes_td3_policies}
\end{figure}

Fig. \ref{fig:results:lidar_regimes_td3_policies} likewise shows the TD3 policies. Unlike in the PPO training schemes, TD3 generally did not benefit from being trained on noisy scenarios and the default policy trained on no-noise map performed the best across all evaluation scenarios.

\subsection{Algorithm Training Comparison (Camera Observation)}\label{section:results:algorithm_training_comparison}
As described in section \ref{subsection:RL}, different RL methods are trained on the \textit{DRL-Robot-Navigation} environment: TD3, PPO, Recurrent PPO (or PPO-LSTM), and DreamerV3. 
\begin{figure}[h]
\centering
\includegraphics[width=\textwidth]{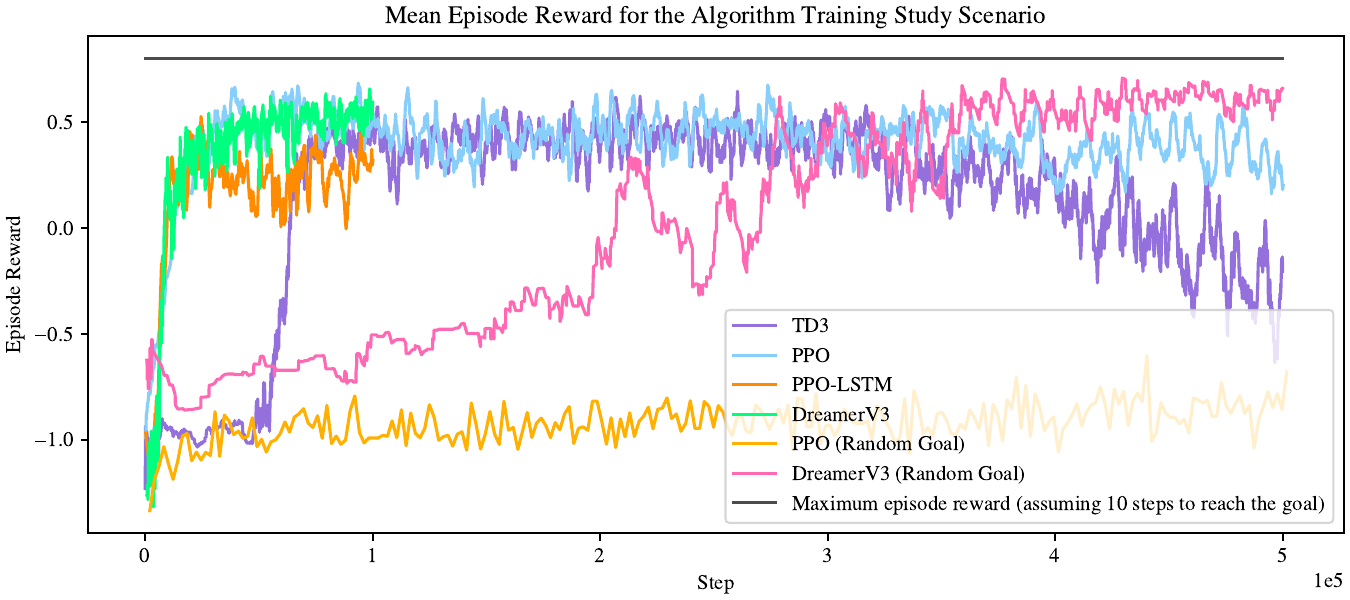}
\caption{Comparison of convergence of different algorithms during training. DreamerV3 was trained with a random goal, while the other models were trained using a static goal in the centre (making DreamerV3s convergence more complicated). }
\label{fig:results:algo_study}
\end{figure}

Fig. \ref{fig:results:algo_study} shows the averaged episode reward during training for the different algorithms, across 500,000 steps of training. 
For the static goal scenario, TD3, PPO, and PPO-LSTM converge to a solution quickly, each one reaching ~0.5 mean reward in less than 100,000 steps. 
The only model that was able to learn to navigate to a random goal was DreamerV3. By comparison, PPO failed to learn to navigate even after 500,000 steps. 
Nonetheless, to better understand the effects of noise on training, a static goal will be used for the camera policies.

\subsection{Camera Navigation with Sensor Denial}\label{section:results:camera_observation_regimes}

\begin{figure}[h]
\centering
\includegraphics[width=\textwidth]{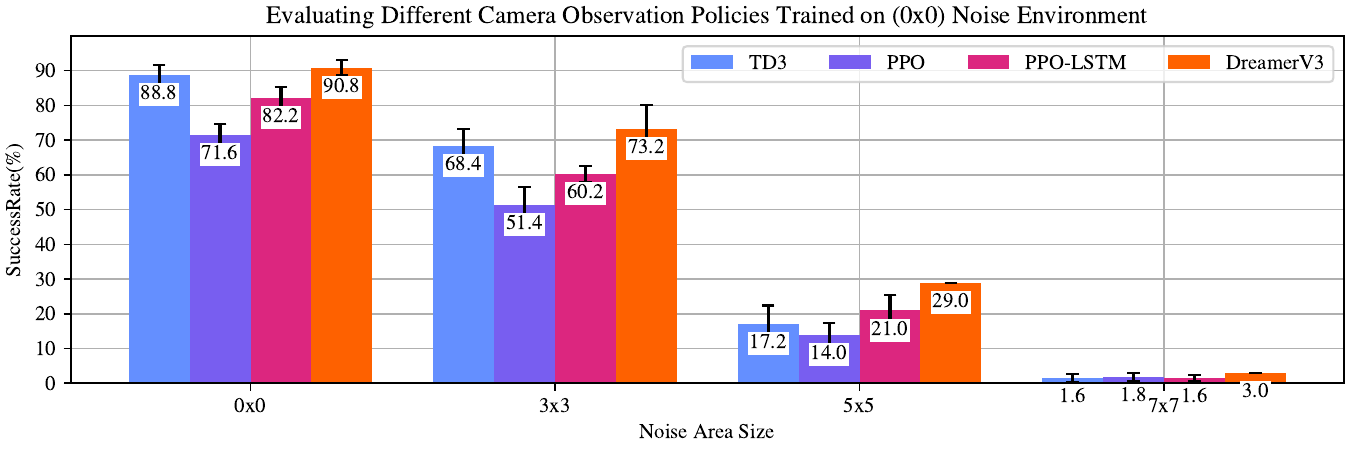}
\caption{Camera noise area study for algorithms trained on the vanilla environment with no sensor denial. The algorithms were then evaluated on the vanilla environment with changing noise area sizes, as shown on the x-axis. The evaluation was performed for 100 episodes and repeated 5 times to establish a standard deviation.}
\label{fig:results:camera_noise_0x0}
\end{figure}

Fig. \ref{fig:results:camera_noise_0x0} shows the evaluation results of the different algorithms, TD3, PPO, PPO-LSTM, and DreamerV3 using the camera observation space. The algorithms are trained on the vanilla 0x0 map, and evaluated on environments of increasing camera denial area: 0x0 (sensor denial), 3x3, 5x5, and 7x7 - referring to the size of the rectangle of the sensor denied area in meters. 
On the 0x0 evaluation scenario, DreamerV3 achieved the highest mean success rate of 90.8\%, with TD3 achieving a similar performance of 88.8\%. PPO-LSTM was third with a mean of 82.2\% and PPO performed the worst with a mean success of 71.6\%. 
On the 3x3 evaluation scenario, the pattern remains the same, with DreamerV3 achieving the highest mean of 73.2\% (with a much larger standard error of 6.8), TD3 close behind with 68.4\%, PPO-LSTM third and PPO last. 
The evaluation scenarios for 5x5 and 7x7 show severe degradation in performance across all the algorithms.

\begin{figure}[h]
\centering
\includegraphics[width=\textwidth]{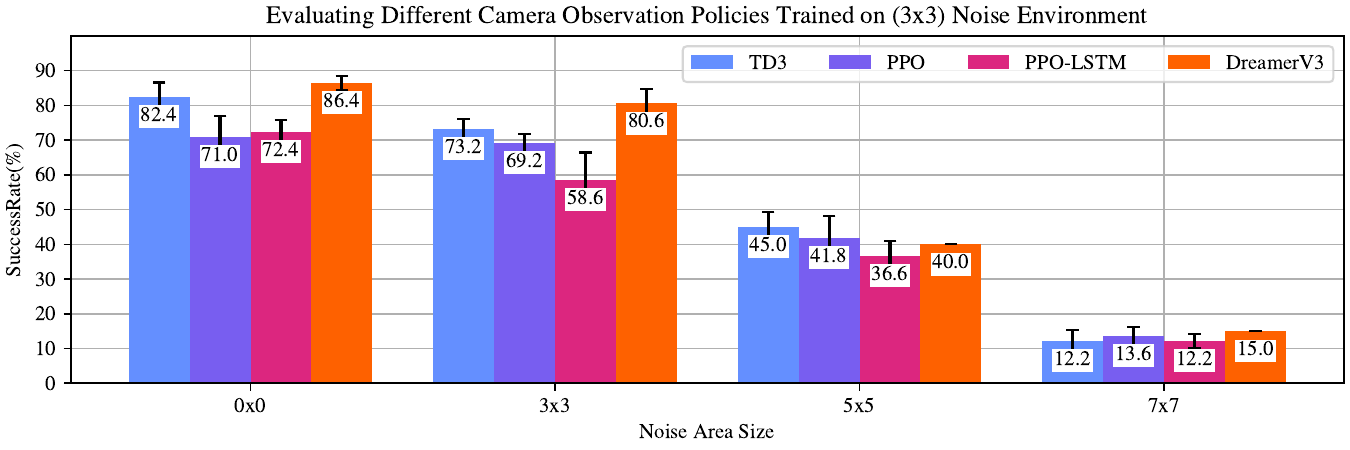}
\caption{Camera noise are study, for algorithms trained on the default environment with 3x3 sensor denial zones. The algorithms were then evaluated on the default environment with changing sensor denial area sizes, as shown on the x-axis. The evaluation was performed for 100 episodes and repeated 5 times to present a mean and standard error.}
\label{fig:results:camera_noise_3x3}
\end{figure}

Similarly, Fig. \ref{fig:results:camera_noise_3x3} shows the evaluation results of the algorithms trained on a map with random 3x3 denied areas. 
When evaluated on the 0x0 scenario, a similar pattern emerges with DreamerV3 achieving the highest mean success rate of 86.4\% (albeit, this is a performance drop compared with 90.8\% when trained on the clean scenario). TD3 is second with 82.4\%, and PPO and PPO-LSTM show the worst performance achieving 71.0\% and 72.4\% respectively. All of the policies show performance drops compared to policies trained on the clean (i.e. no sensor denied areas) environment. 

When evaluated on the 3x3 scenario, DreamerV3 performs the best achieving 80.6\% mean success rate. This is an improvement from the vanilla model (from 73.2\%). TD3 mean success rate is 73.2\%, PPO 69.2\%, and PPO-LSTM is last with 58.6\%. Likewise, TD3 and PPO also show improvements from the vanilla models. PPO-LSTM is the only model that did not report an improvement from the vanilla model.  

All models report an improvement in evaluation performance on 5x5 and 7x7 noise area maps (compared with models trained on no noise in the environment), showing that training on some noise in the scenario generally improves the evaluation performance on noisy evaluation scenarios. Still, these improvements are not so significant that the models are reliably achieving the goal. The downside of training the model on the 3x3 scenario with sensor-denied areas is that these models appear to suffer from performance degradation when evaluated on the 0x0 scenario.

\subsection{Camera Navigation with Sensor Denial (No Obstacle Map)}\label{section:results:navigation_noise_camera_no_obstacle}
Section \ref{section:results:camera_observation_regimes} shows that the training regimes have an effect on policy behaviour in sensor-denied areas. To better understand this behaviour, they are evaluated on the simplified, no obstacle map, as described in section \ref{subsection:evaluation}.

\begin{figure}[h]
\centering
\includegraphics[width=\textwidth]{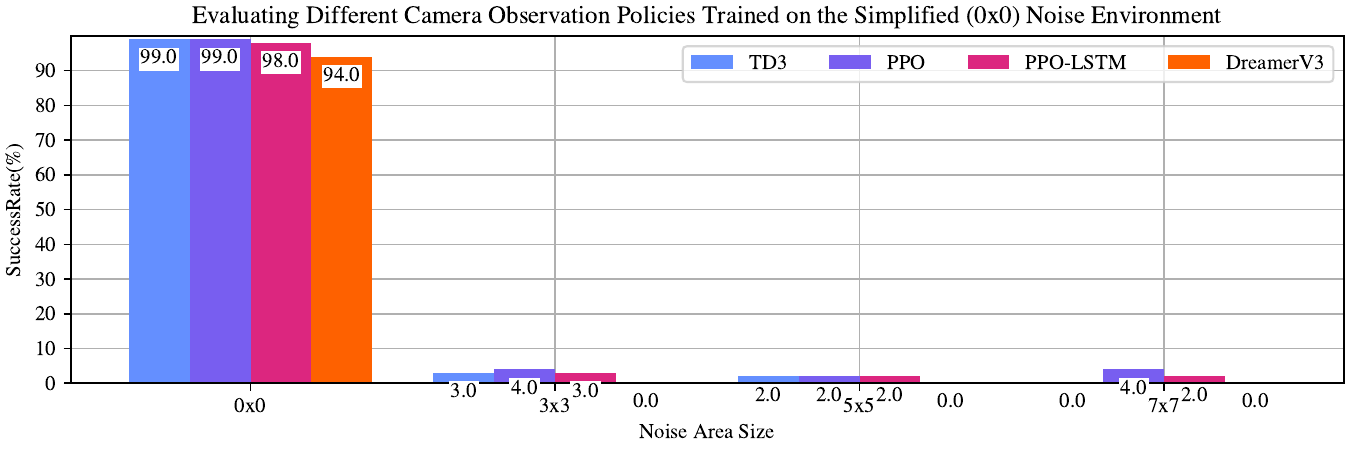}
\caption{Camera noise area study for algorithms trained on the default environment with 0x0 noise. The algorithms were then evaluated on the simplified map with changing noise area sizes, as shown on the x-axis. The evaluation was performed for 100 episodes.}
\label{fig:results:camera_noise_0x0_no_obstacle}
\end{figure}

Fig. \ref{fig:results:camera_noise_0x0_no_obstacle} shows the evaluation results of the different algorithms, TD3, PPO, PPO-LSTM, and DreamerV3 using the camera observation space. The algorithms are trained on the vanilla 0x0 (no sensor denial) default map, and evaluated on the simplified, no obstacle map, with variations of increasing, static camera denial area which is located in the centre of the map: 0x0 (no sensor denial), 3x3, 5x5, and 7x7 - referring to the size of the rectangle of the sensor denied area in meters. 
All four of the models exhibit similar behaviour of successfully navigating in nearly all of the episodes on the 0x0 evaluation, but then consistently failing when any noise is introduced (with a few successes likely caused by spawning near the goal). 

\begin{figure}[h]
\centering
\includegraphics[width=\textwidth]{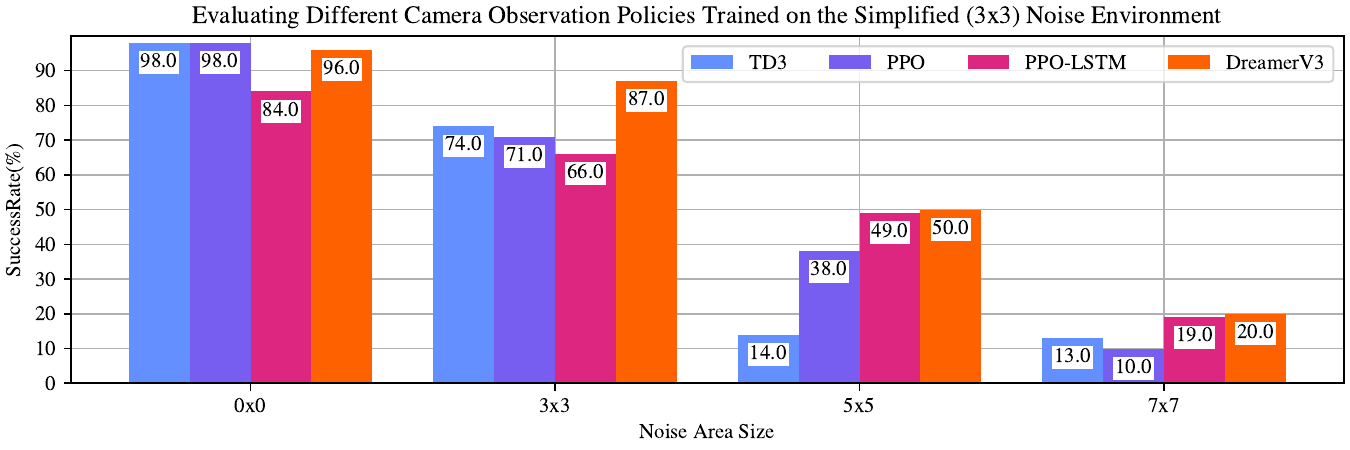}
\caption{Camera noise are study, for algorithms trained on the default environment with 3x3 noise. The algorithms were then evaluated on the simplified map with changing noise area sizes, as shown on the x-axis. The evaluation was performed for 100 episodes.}
\label{fig:results:camera_noise_3x3_no_obstacle}
\end{figure}

Similarly, Fig. \ref{fig:results:camera_noise_3x3_no_obstacle} shows the evaluation results of the different algorithms trained on the default map with random 3x3 noise areas. The 0x0 evaluation does not significantly change, with TD3, PPO, and DreamerV3 still successfully navigating to the target 98, 98, and 96 times out of 100. 
For the 3x3 evaluation, TD3 reached the goal 74 times, PPO 71 times, PPO-LSTM 66 times, and DreamerV3 showed the best performance with 87 times. This is a significant improvement over the models trained on the 0x0 (no-noise) environment - a sign that the policies have learned to navigate to the goal with noise present. The 3x3-trained models also show performance improvements over the 0x0-trained models when evaluated on 5x5 and 7x7 scenarios. However, they do not achieve high (90+) success rates, which would be an indication of the policies truly solving the scenarios.

\begin{figure}[h]
\centering
\includegraphics[width=\textwidth]{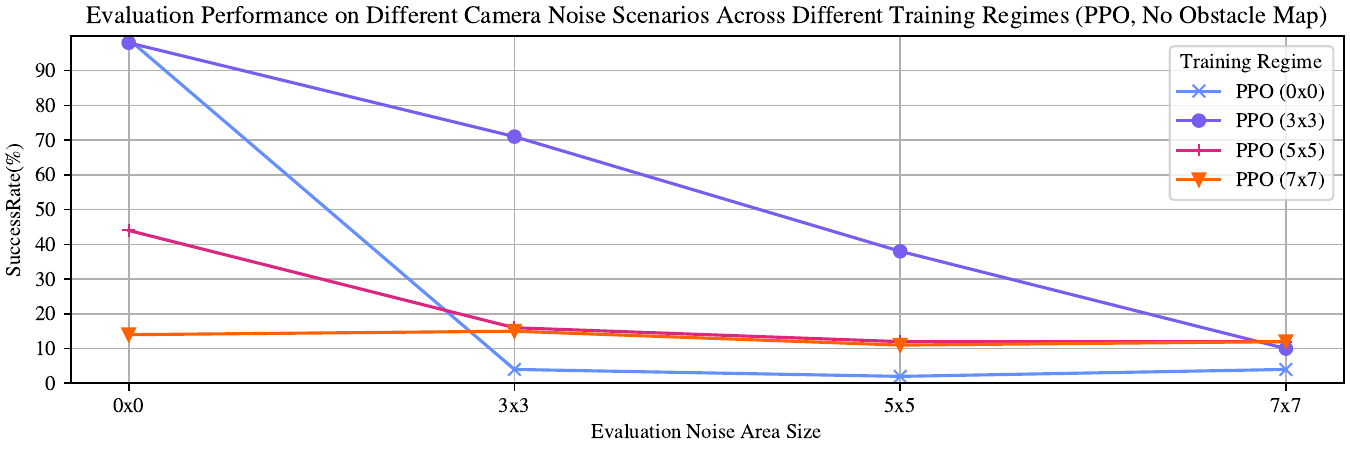}
\caption{PPO policies trained on different noise regimes evaluated on the simplified map with varying noise area sizes.}
\label{fig:results:ppo_policies_no_obstacle}
\end{figure}

Fig. \ref{fig:results:ppo_policies_no_obstacle} shows the PPO policies trained on different camera noise regimes evaluated on the no-obstacle map. 
It shows that the PPO (0x0) and PPO (3x3) perform the best when evaluated on the 0x0 scenario, achieving 99 and 98 successful episodes. The PPO (0x0) fails to navigate to the goal when evaluated on any noise maps. PPO (5x5) and PPO (7x7) both failed to navigate successfully in the 0x0 evaluation, a sign that too much noise hindered the learning of those models.

\begin{figure}
    \subfigure[\label{fig:results:paths_ppo_0x0_eval_0x0_camera}]{\includegraphics[width=.49\textwidth]{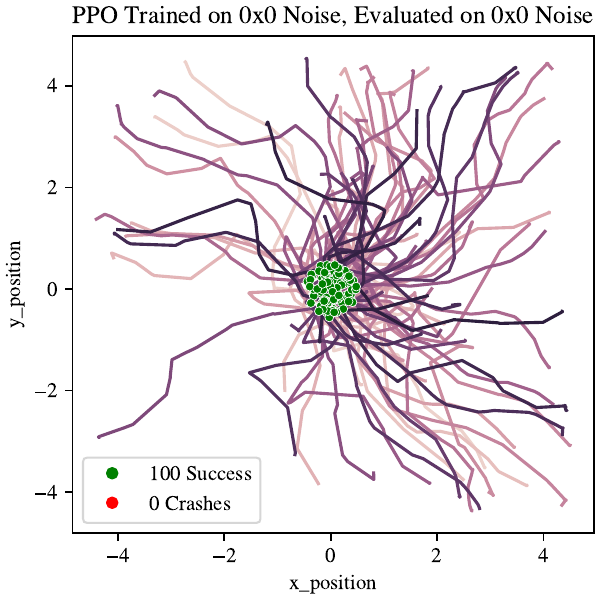}}%
    \quad  
    \subfigure[\label{fig:results:paths_ppo_0x0_eval_3x3_camera}]{\includegraphics[width=.49\textwidth]{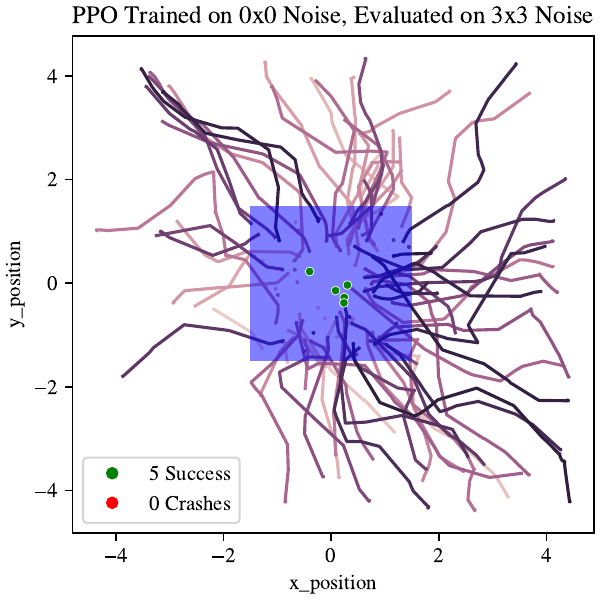}}%
    \quad  
    \subfigure[\label{fig:results:paths_ppo_3x3_eval_0x0_camera}]{\includegraphics[width=.49\textwidth]{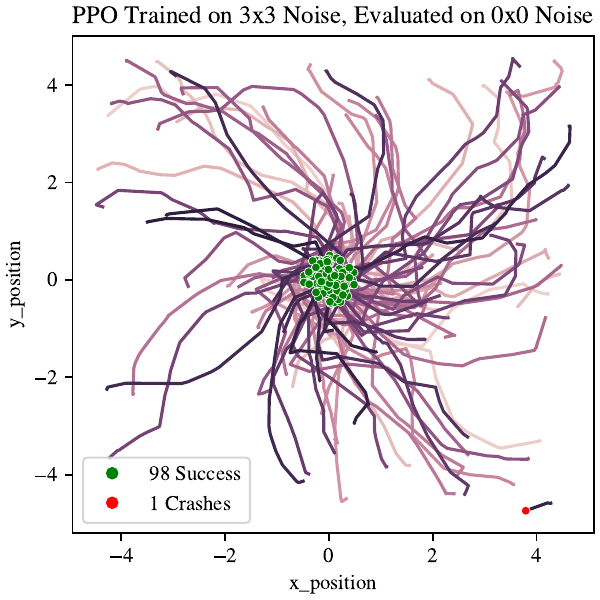}}%
    \quad  
    \subfigure[\label{fig:results:paths_ppo_3x3_eval_3x3_camera}]{\includegraphics[width=.49\textwidth]{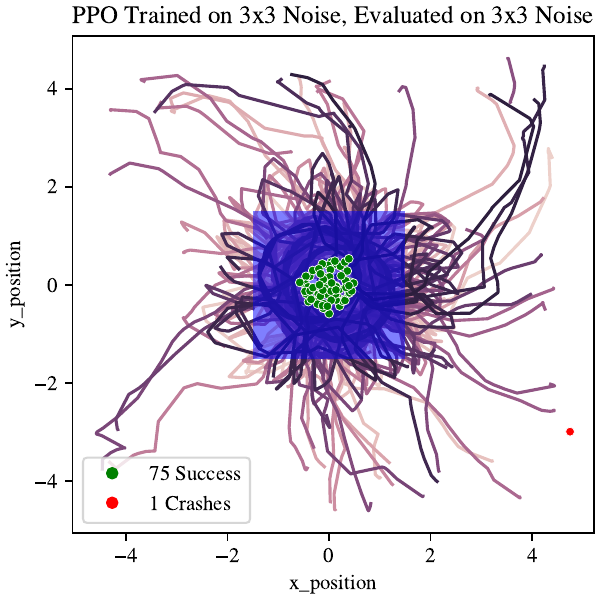}}%
     \caption{100 path plots generated during evaluation by vanilla policy (PPO 0x0) evaluated on 0x0 map \textbf{(a)}, vanilla policy (PPO 0x0) evaluated on 3x3 sensor denial map \textbf{(b)}, adversarially trained (PPO 3x3) evaluated on 0x0 sensor denial map \textbf{(c)}, and adversarially trained (PPO 3x3) evaluated on 3x3 sensor denial map \textbf{(d)}. Red dot shows a collision, green dot shows a successful navigation to the goal. Blue zone shows the area of sensor denial. }
    \label{fig:results:qualitative_camera_ppo_regime_runs}
\end{figure}

Fig. \ref{fig:results:qualitative_camera_ppo_regime_runs} shows the 100 path plots generated during the evaluation of 2 different policies (PPO 0x0 and PPO 3x3) on two different evaluation maps (0x0 and 3x3 static sensor denial zones, static goal in the centre, and no obstacles present). Fig. \ref{fig:results:paths_ppo_0x0_eval_0x0_camera} shows the vanilla PPO evaluated on the 0x0 map. The same vanilla policy is evaluated on a map with a static 3x3 noise area around the goal in Fig. \ref{fig:results:paths_ppo_0x0_eval_3x3_camera}. The policy is unable to navigate to the goal and shows a significant decrease in performance (from 100 successes to 5 successes). When the robot enters the noise area, the camera fails, and the robot stops moving. This can be seen as many paths end inside the blue noise zone and do not continue.

In contrast, when adversarially trained with 3x3 random areas present in the environment and evaluated on the 3x3 map as shown in Fig. \ref{fig:results:paths_ppo_3x3_eval_3x3_camera}, the number of successful episodes increases to 75. When inside the blue sensor denied area, the agent simply moves in circles, as can be seen by path traces in the blue zone. This is not a reliable way of navigating with obstacles present in the environment but leads to success when no obstacles are present in the zone of sensor denial. The policy exploits the reward structure of the scenario, opting for a high-risk strategy of moving through the sensor denied areas. 
Hence, like the Lidar examples shown in Fig. \ref{fig:results:qualitative_lidar_ppo_regime_runs}, the policies have learned a high-risk strategy of blindly navigating when sensor readings are uncertain. 

Since we are evaluating on the simplified map, no obstacles are present. Clearly, once in a sensor-denied area, the policies have no way of avoiding obstacles and this is deliberately not reflected in these results, in order to understand the policy behaviour in sensor-denied zones. Still, because the models were trained on the default map, they have learned to take the risky actions when in sensor-denied zones with obstacles present. This likely means that due to the reward shaping of the environment, it was more rewarding for the policy to take the high risk actions. However this only holds up with small amounts of sensor denied zones in the environment, as shown in Fig. \ref{fig:results:ppo_policies_no_obstacle}, as larger amounts of noise result in failure to learn basic navigation behaviour.

\subsection{Discussion}\label{section:results:discussion}
The overall best performer for the camera navigation was DreamerV3. DreamerV3 consistently outperformed other models for the camera observations for both 0x0 and 3x3 noise area evaluations, when trained on both 0x0 and 3x3 environments (Fig. \ref{fig:results:camera_noise_0x0} and \ref{fig:results:camera_noise_3x3}). This might result from the DreamerV3 having a different structure to the other models and building a world model due to the autoencoder structure.
We hypothesized that it might be able to overcome the sensor denied areas due to the autoencoder/recurrent structure. While it did outperform other models when trained and evaluated on the 3x3 environment (it reached a mean of 80.6 \% in Fig. \ref{fig:results:camera_noise_0x0}) that still does not match the baseline performance of 90.8\% when trained and evaluated on the no noise environments. 
DreamerV3 was also the only model to have learned to navigate to a random goal using the camera-only observation space within 500,000 steps, as shown in section \ref{section:results:algorithm_training_comparison}. Other models either required the Lidar policy (which contained information about distance and angle to target), or required a static goal to be able to converge to a solution using the camera observation space. 
DreamerV3 was also convenient to use as it did not require any hyperparameter tuning.
The best performer (between TD3 and PPO) for the Lidar modality was TD3. The 0x0 trained TD3 model generally outperformed most variations on PPO across different evaluations.

In sections \ref{section:results:lidar_observation_quantitative} and \ref{section:results:navigation_noise_camera_no_obstacle} we have learned that training policies with some noise present in the environment present an interesting result: these policies are more successful at navigating to the goal, but also take more high-risk actions when sensor readings are uncertain or completely denied, which may lead to colliding with obstacles more often - a terminating event in our scenario. 
This is a limitation of training RL policies for safety-critical scenarios. Training them to maximize the reward and reach the goal in uncertain scenarios may lead to the policies learning to select actions that are deemed rewarding and may lead to achieving a reward, even if these actions may lead to a catastrophic scenario. 

Korkmaz \cite{korkmaz_adversarial_2023} reports that adversarially trained policies are more vulnerable to shifts in e.g. brightness or blurring than vanilla trained ones. While we did not consider adversarial learning algorithms (Korkmaz used SA-DDQN which is based on the SA-MDP and RADIAL algorithms, while we only trained our models on environments with adversarial perturbations), this could be mapped to our problem in that our noise-trained policies (e.g. 3x3 PPO) result in a larger amount of crashes and are less robust on the 0x0 evaluation, compared with vanilla PPO (0x0).

Mirowski et al. \cite{mirowski_learning_2017} perform a similar study which includes static mazes and large random mazes. These are equivalent to static goal (that we performed using camera), and random goal (that we performed using Lidar).
In their paper, algorithms that used recurrence outperformed non-recurrent algorithms. We did not see such a trend in our results, and PPO-LSTM generally performed poorly (ignoring DreamerV3 which is also recurrent). This is surprising given that navigation is a partially observable Markov decision process (MDP), and hence, policies should benefit from memory.

\section{Conclusion and Further Work}\label{section:Conclusion}

This study has shown quantitative results of how common reinforcement learning algorithms perform with different modalities - camera and Lidar - on the \textit{DRL-Robot-Navigation} environment, and the different effects of training and evaluating on environments containing noisy and sensor denied areas. 
We observed that training DRL models on noisy environments creates policies that, generally perform better on noisy environments, compared with models trained on vanilla, no noise environments. However, we found that these policies opted for riskier actions when presented with a noisy or faulty observation: ones that might result in successful navigation to the goal, but also in collision with another object. 
While these models result in better performance by quantitative metrics, this behaviour is unacceptable from the point of view of safety-critical scenarios and shows a need for more robust RL methods that are resilient to noise or sensor outages in the scenarios. 
Hence, we open-source the environments that we built on top of the \textit{DRL-Robot-Navigation} to enable other researchers to evaluate their model performance on these scenarios. 
The results presented here should serve as a benchmark for any future studies on strategies of how to deal with faulty sensors.
While we present evaluation for Gaussian noise to Lidar and camera outages, different perturbations also should be integrated (e.g. \cite{korkmaz_understanding_2024} investigates transformations such as compression artefacts, brightness and contrast, and blur for camera observations). 
We consider that work on autoencoders could lead to identifying sensor failures, and in potentially serving as a backup that is able to fill out the gaps in case of sensor outages. 
Sensor fusion strategies should also be investigated.
Hierarchical reinforcement learning could lead to better performance (as proposed by Havens et al. \cite{havens_online_2018}) as it allows for the top hierarchy to detect adversaries. This could lead to better performance in tasks with noise, as the local goals might change through the course of the episode - from navigating to a target to getting out of noise. 

Finally, we suggest further developments to the environment, to include environmental interactions, along with more elaborate models of sensor attacks.

\begin{appendices}
\section{Learning Rate Sensitivity}
As the algorithms (apart from DreamerV3) are sensitive to learning rate, we initially perform a learning rate study - we then continue using this learning rate throughout.

\begin{figure}[h]
\centering
\includegraphics[width=\textwidth]{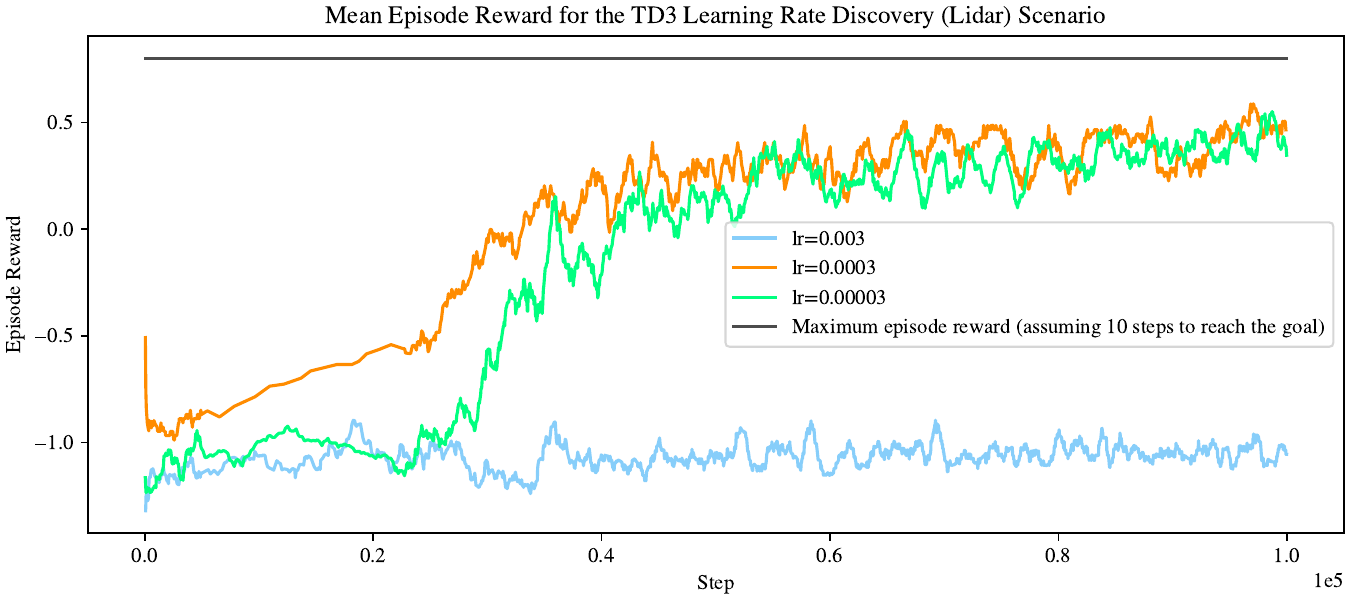}
\caption{TD3 Learning Rate Discovery (Lidar).}
\label{fig:appendix:td3_lr}
\end{figure}
Fig. \ref{fig:appendix:td3_lr} shows the learning rate discovery for the TD3 algorithm using the Lidar observation space. Both, 0.0003 and 0.00003 values converge to approximately a max of 0.5 episode reward, but 0.0003 appears to converge more quickly. 

\begin{figure}[h]
\centering
\includegraphics[width=\textwidth]{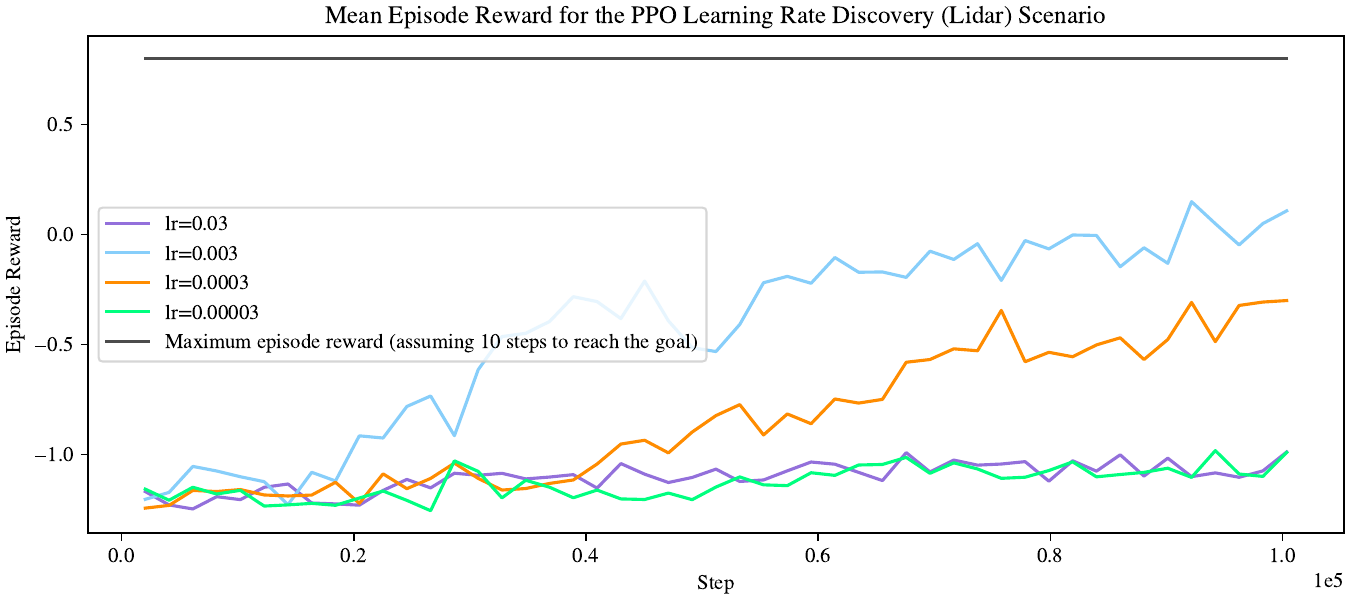}
\caption{PPO Learning Rate Discovery (Lidar).}
\label{fig:appendix:ppo_lr}
\end{figure}
Fig. \ref{fig:appendix:ppo_lr} shows the learning rate discovery for the PPO algorithm using the Lidar observation space. A value of 0.003 appears to be the optimal value converging quicker than 0.0003.

\begin{figure}[h]
\centering
\includegraphics[width=\textwidth]{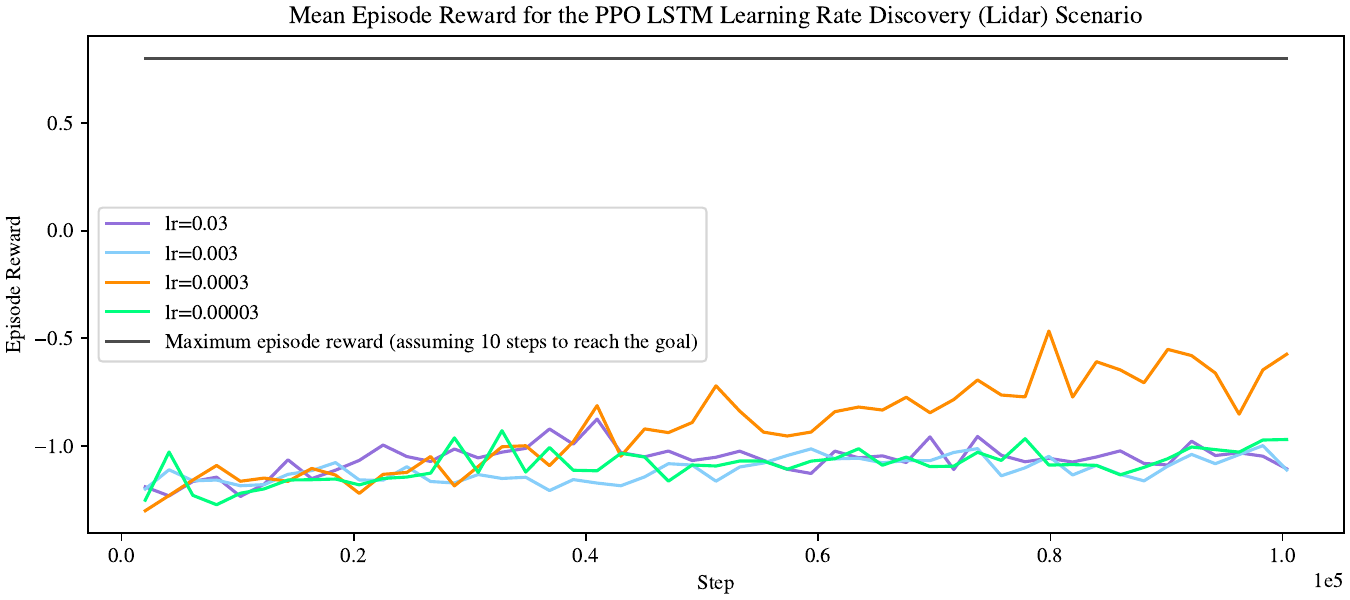}
\caption{PPO LSTM Learning Rate Discovery (Lidar).}
\label{fig:appendix:ppo_lstm_lr}
\end{figure}
Fig. \ref{fig:appendix:ppo_lstm_lr} shows the learning rate discovery for the PPO LSTM algorithm using the Lidar observation space. No clear convergence was achieved - this is surprising because we expected recurrence networks to work on this problem. PPO without recurrence performs much better with the Lidar observation space as shown in Fig. \ref{fig:appendix:ppo_lr}.


\begin{figure}[h]
\centering
\includegraphics[width=\textwidth]{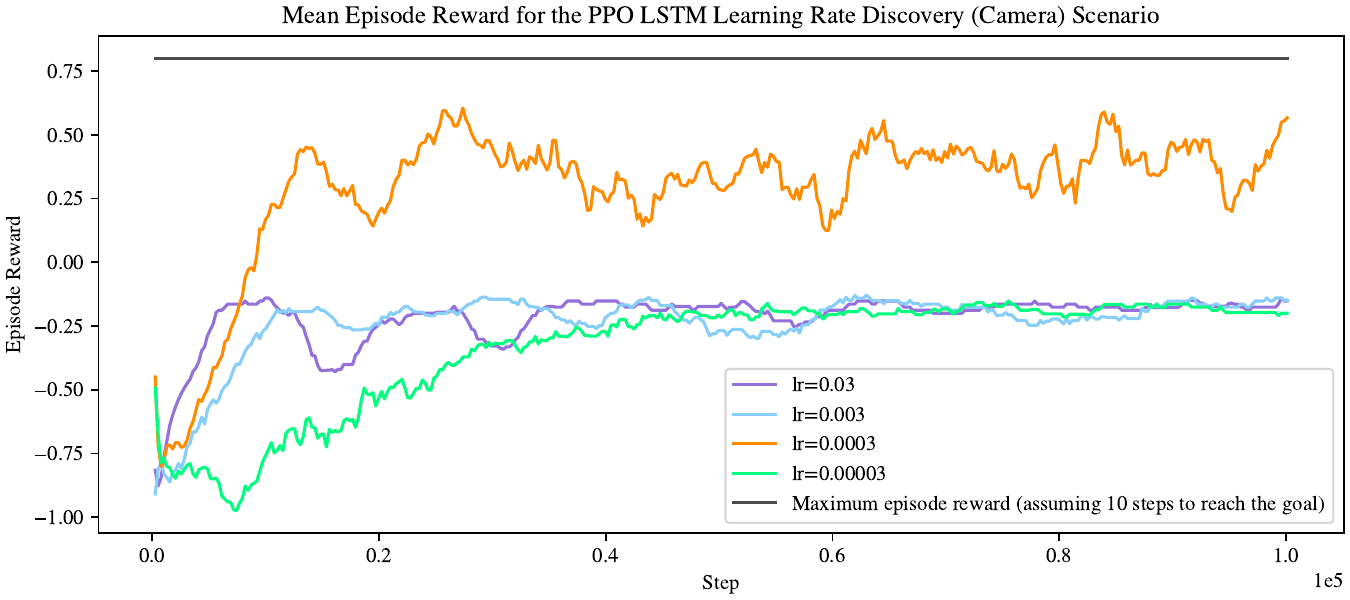}
\caption{PPO LSTM Learning Rate Discovery (Camera).}
\label{fig:ppo_lstm_cam_lr}
\end{figure}

Fig. \ref{fig:ppo_lstm_cam_lr} shows the learning rate discovery for the PPO LSTM algorithm using the camera observation space. A value of 0.0003 is the only one to converge to a solution. We will use this value of 0.0003 for both PPO and PPO LSTM camera runs.

\end{appendices}

\backmatter

\bmhead{Supplementary information}
The modified \textit{DRL-Robot-Navigation} environment is available \url{https://github.com/mazqtpopx/cranfield-navigation-gym}.

\bmhead{Acknowledgements}
This work with MW and PC is supported by Leonardo UK with Cranfield University, as well as WG and AT is supported by EPSRC TAS-S: Trustworthy Autonomous Systems: Security (EP/V026763/1).  

\section*{Declarations}
\begin{itemize}
\item Funding: This work with MW and PC is supported by Leonardo UK with Cranfield University, as well as WG and AT is supported by EPSRC TAS-S: Trustworthy Autonomous Systems: Security (EP/V026763/1).  
\item Conflict of interest/Competing interests N/A
\item Ethics approval and consent to participate: N/A
\item Consent for publication: N/A
\item Data availability: N/A
\item Materials availability: N/A
\item Code availability: code is available \url{https://github.com/mazqtpopx/cranfield-navigation-gym}.
\item Author contribution: All authors contributed to the study conception and design. Material preparation, data collection and analysis were performed by MW, and PC. The first draft of the manuscript was written by MW and all authors commented on previous versions of the manuscript. All authors read and approved the final manuscript.

\end{itemize}
\bibliography{mw-references.bib, weisi-ref.bib}
\end{document}